\documentclass[lettersize, journal]{IEEEtran}

\usepackage[caption=false,font=footnotesize,labelfont=sf,textfont=sf]{subfig}
\usepackage{array}
\usepackage{textcomp}
\usepackage{stfloats}
\usepackage{url}
\usepackage{verbatim}
\usepackage{graphicx}
\usepackage{cite}
\usepackage{times} 
\usepackage{amsmath, amsfonts} 
\usepackage{amssymb}  
\usepackage{enumerate}
\usepackage{bm}
\usepackage{cases}
\usepackage{mathtools}
\usepackage{multirow}
\usepackage{colortbl}
\usepackage{booktabs} 
\usepackage{makecell}
\usepackage{algpseudocode}
\usepackage{algorithmicx}
\usepackage{algorithm}
\usepackage{ragged2e}
\usepackage{tabularx}
\usepackage{xcolor}
\usepackage[para]{threeparttable}
\usepackage{colortbl}
\usepackage[hidelinks]{hyperref} 
\newcolumntype{C}{>{\Centering\arraybackslash}X}
\newcolumntype{L}{>{\raggedright\arraybackslash}X}
\newcolumntype{R}{>{\raggedleft\arraybackslash}X}

\begin{document}
\title{\LARGE \bf
IR-STP: Enhancing Autonomous Driving with Interaction Reasoning in Spatio-Temporal Planning 

\author{Yingbing Chen$^{1}$, Jie Cheng$^{1}$, Lu Gan$^{1}$, Sheng Wang$^{1}$, Hongji Liu$^{1}$, Xiaodong Mei$^{1}$, and Ming Liu$^{1,2}$}

\thanks{
    Manuscript received Aug. 15, 2023; revised Nov. 23, 2023; accepted Jan. 14, 2024. This work was supported by Lotus Technology Ltd. under cooperation project (R00082) with Hongkong University of Science and Technology (GZ), and Project of Hetao Shenzhen-Hong Kong Science and Technology Innovation Cooperation Zone(HZQB-KCZYB-2020083), awarded to Prof. Ming Liu.  (\textit{Corresponding author: Ming Liu.}).
}


\thanks{$^{1}$Yingbing Chen, Jie Cheng, Lu Gan, Sheng Wang, Hongji Liu, and Xiaodong Mei are with The Hong Kong University of Science and Technology, Hong Kong SAR, China. {\tt\footnotesize \{ychengz, jchengai, lganaa, swangei, hliucq, xmeiab\}@connect.ust.hk.}}%

\thanks{$^{2}$Ming Liu is with the HKUST (GZ), Guangzhou, China, and with the HKUST Shenzhen-Hong Kong Collaborative Innovation Research Institute, Futian, Shenzhen, and also with the Unity Drive Innovation Technology Co. Ltd., Shenzhen, China. {\tt\footnotesize eelium@hkust-gz.edu.cn.}}%

}

\markboth{IEEE TRANSACTIONS ON INTELLIGENT TRANSPORTATION SYSTEMS}%
{Chen \MakeLowercase{\textit{et al.}}: Enhancing Autonomous Driving with Interaction Reasoning}

\IEEEpubid{0000--0000/00\$00.00~\copyright~2024 IEEE}
\IEEEpubidadjcol 

\maketitle

\begin{abstract}
    Considerable research efforts have been devoted to the development of motion planning algorithms, which form a cornerstone of the autonomous driving system (ADS). Nonetheless, acquiring an interactive and secure trajectory for the ADS remains challenging due to the complex nature of interaction modeling in planning. Modern planning methods still employ a uniform treatment of prediction outcomes and solely rely on collision-avoidance strategies, leading to suboptimal planning performance. To address this limitation, this paper presents a novel prediction-based interactive planning framework for autonomous driving.
    Our method incorporates interaction reasoning into spatio-temporal (s-t) planning by defining interaction conditions and constraints. Specifically, it records and continually updates interaction relations for each planned state throughout the forward search. 
    We assess the performance of our approach alongside state-of-the-art methods in the CommonRoad environment. 
    Our experiments include a total of 232 scenarios, with variations in the accuracy of prediction outcomes, modality, and degrees of planner aggressiveness. The experimental findings demonstrate the effectiveness and robustness of our method. It leads to a reduction of collision times by approximately 17.6\% in 3-modal scenarios, along with improvements of nearly 7.6\% in distance completeness and 31.7\% in the fail rate in single-modal scenarios. For the community's reference, our code is accessible at \href{https://github.com/ChenYingbing/IR-STP-Planner}{\color{blue}{https://github.com/ChenYingbing/IR-STP-Planner}}.

\end{abstract}

\renewcommand{\arraystretch}{1.0}


\vspace{-0.5em}
\section{Introduction\label{sect:intro}}

\IEEEPARstart{A}{chieving} effective and safe trajectory planning for autonomous vehicles (AVs) necessitates the integration of interaction modeling into the motion planning algorithm, as it enhances the adaptability and safety of AVs in intricate traffic environments. 
One of the predominant planning methodologies is prediction-based planning. When addressing probabilistic prediction outcomes, certain investigations \cite{zhang2021hierarchical, fu2021trajectory} have delved into search-based approaches aimed at generating smooth and low-risk trajectories for AVs. Meanwhile, other approaches \cite{gao2018online, ding2019safe} have introduced techniques for smoothing trajectories by utilizing collision-free driving corridors extracted from deterministic prediction results. Furthermore, fail-safe \cite{pek2020tro_fail_safe} and contingency \cite{cui2021lookout} planners concentrate on planning approaches reliant on multi-modal prediction results. Their objective is to ensure robust and secure navigation performance for AVs across diverse prospective scenarios. Nonetheless, these methodologies exhibit a reactive nature. They neglect the integration of interaction modeling and primarily rely on evading ``collisions'' or ``collision risks'' inferred from prediction outcomes. This can lead to less effective planning performance due to their insufficient consideration of interactions.

\begin{figure}[!t]
    \centering
    \includegraphics[width=0.9\linewidth]{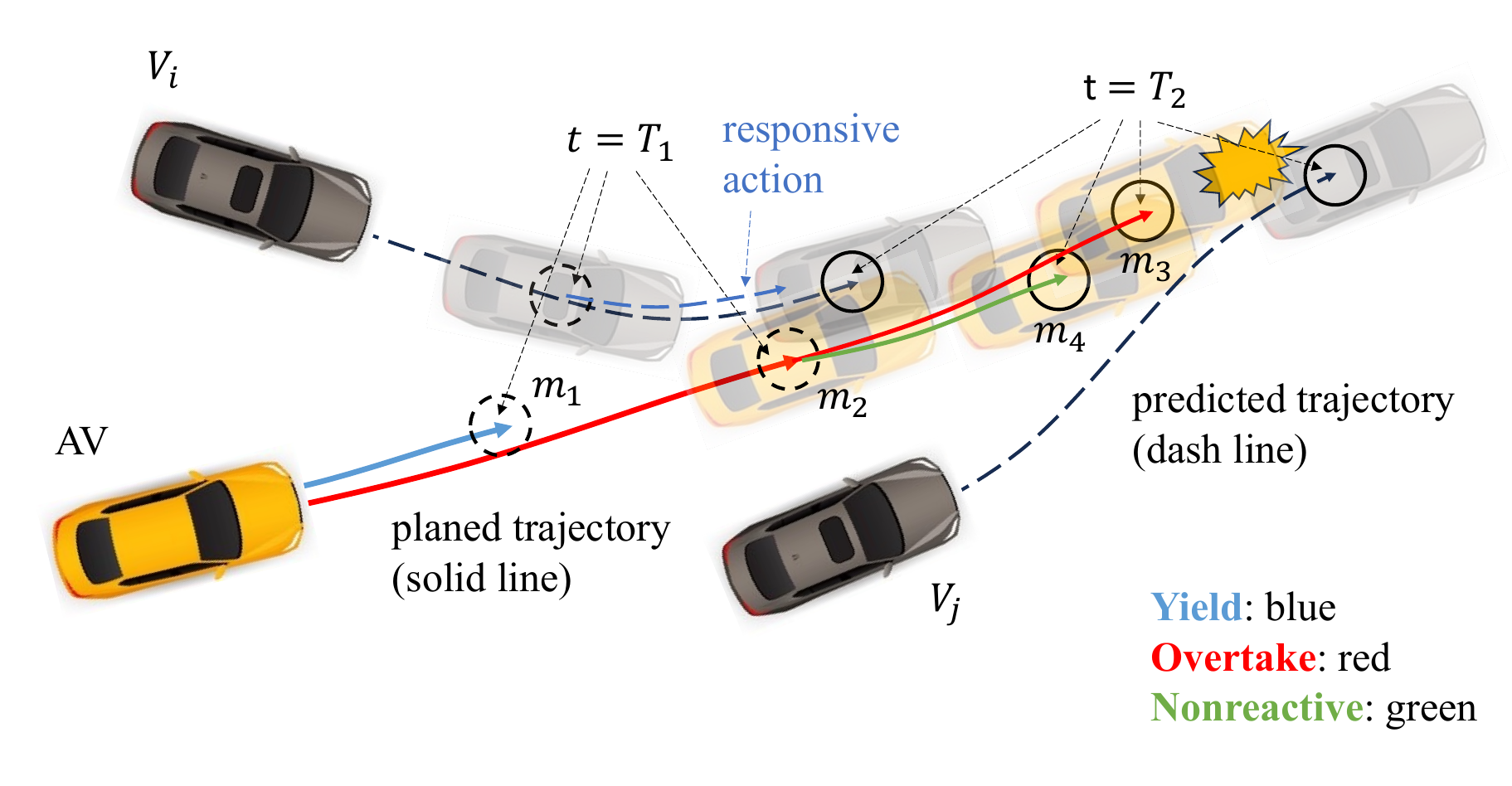}\vspace{-1.0em}
    \caption{Example scenario: The motion planner is tasked with generating trajectories that dynamically interact  with the predicted future trajectories of agents $V_i$ and $V_j$. Here, $T_2$ signifies the planning horizon, and $m_{(\cdot)}$ denotes the plausible future motions of the AV at time stamps $T_1$ and $T_2$. Multiple trajectory solutions are available. $m_1$: The most cautious choice, entailing yielding to both $V_i$ and $V_j$ to ensure safety. $m_2$-$m_3$: A purely reactive trajectory which reacts to $V_i$'s trajectory at all time steps, but leads to a collision with $V_j$'s trajectory at the $T_2$ time point. $m_2$-$m_4$: An interactive and more efficient trajectory strategy. Initially, it reacts to $V_i$'s trajectory to prevent collisions within the time interval $T_1$, reaching position $m_2$ at that moment. Then, it remains nonreactive to $V_i$'s subsequent motions ($t \in (T_1, T_2]$), assuming that $V_i$ will react to the AV's motion. Eventually, the trajectory reaches $m_4$ at time $T_2$, safely accommodating $V_j$'s trajectory, while $V_i$ takes responsive action to evade collisions with the AV.}
    \label{fig:intro_overview}
\vspace{-2.0em}
\end{figure}

\IEEEpubidadjcol 

Specifically, current prediction-based planner (e.g., \cite{li2022motion, ma2022alternating}), which lack interaction reasoning, face challenges in successfully devising plans for scenarios similar to the illustration presented in Fig. \ref{fig:intro_overview}.
Relying exclusively on a reactive trajectory in such scenarios brings forth several limitations:
i) A pure collision avoidance strategy struggles to capture the intricate behavioral patterns exhibited by traffic agents during interactions.
ii) This necessitates precise predictions of the movements of other traffic agents under all circumstances, a task that is frequently impractical in real-world scenarios.
iii) The solution space can quickly become inundated with predicted trajectories, especially in densely congested traffic scenarios, potentially leading to a dearth of feasible solutions for the AV.
These limitations primarily arise from their uniform treatment of prediction outcomes, which fails to consider the distinct roles played by traffic agents and their trajectories during interactions. For instance, responding to an inaccurate predicted trajectory during yielding and overtaking could lead to differing effects on navigation performance. 
Furthermore, as depicted in Fig. \ref{fig:intro_overview}, a thoughtfully designed interaction model can treat these prediction outcomes in a disparate manner, resulting in a substantial enhancement of planning performance (as demonstrated by the presented interactive trajectory). Building upon these insights, our paper primarily concentrates on the integration of such interaction modeling into the motion planning algorithm.


In this study, we present an interactive planning approach that explicitly incorporates interaction modeling into its methodology. To achieve this, we enhance the widely adopted s-t search frameworks \cite{cheng2022real, li2021spatial}, which were initially designed without interaction modeling. 
Our approach integrates interaction modeling into s-t planning by recording and updating interaction relationships for every planned state throughout the planning process. More specifically, the interaction relationships are modified through forward search, guided by the interaction formulations we put forth. These formulations clarify appropriate interaction dynamics, encompassing reactions (e.g., overtaking and yielding), and delineate strategies to influence the trajectories of other agents, providing guidance on the timing and execution of these actions. The primary contributions of this study encompass:


\begin{itemize}
    \item Formulating and mathematically modeling traffic interactions, coupled with developing strategies for maintaining and updating them within a spatio-temporal searcher.

    \item A novel interactive planning framework for AVs, which demonstrates efficacy and robustness in diverse simulation settings and prediction conditions.

    \item Comprehensive open-source prediction-based navigation system for autonomous driving, accompanied by its integration into a closed-loop traffic simulation platform.
\end{itemize}


\section{Related Work}

Motion planning, an essential component for autonomous vehicles (AVs) and applications \cite{ javanmardi2023sdn, robinson2023autodetect} in intelligent transportation systems, can be classified into two primary categories: reactive planners and interactive planners, based on their consideration of potential reactions from other agents towards AVs. Moreover, motion planners can be further divided into classical methods and learning-based methods, depending on their reliance on machine learning technologies.

\vspace{-0.5em}
\subsection{Classical Methods}

Collision avoidance\cite{singh2022optimizing, cheng2022real} and collision risk minimization\cite{wang2021decision, fu2023efficient} were the predominant classical methods for reactive planners. Nevertheless, risk-averse approaches frequently yielded conservative outcomes. Additionally, these approaches faced challenges in capturing the intricate dynamics between traffic agents due to the inadequate consideration of elements like cooperative and competitive relationships among them. Despite claims of being interactive planners, multi-modal works like contingency planning \cite{cui2021lookout} and branch-MPC \cite{chen2022interactive} maintained a risk-neutral stance. They primarily focused on minimizing anticipated costs across the scenario tree, with limited consideration of interactions in their methodologies.

In contrast, interactive planners extended their approach beyond mere collision avoidance by integrating the responses of other agents into their methodologies. For instance, game theory-based methods \cite{tian2020game, chandra2022gameplan} were recognized for their considerable computational complexity and relied on the assumption that all traffic agents followed the predefined game rules. Another strategy involved the utilization of MDP or POMDP-based techniques \cite{zhang2020efficient, luo2021interactive}, entailing the direct simulation of potential interactions within the entire traffic milieu. However, accurately simulating complex traffic interactions resulted in significant computational costs, as uncertainties increased exponentially with the planning horizon.
Moreover, MDP-based methods required detailed modeling of traffic agents, which posed a challenge due to the complex interplay between agents' behaviors and environmental factors that are not easily modeled.
In comparison, our approach operates as an interactive planner based on upstream prediction outcomes. By considering prediction results, we are able to avoid the need for complex traffic simulations and maintain the flexibility to adjust interaction roles based on specific relation formulations at different time steps. This ability to adapt to time-varying interaction relations is a crucial advantage of our approach when compared to other interactive methods.

\begin{table}[!t]
\centering
\caption{Comparisons of existing interactive planners}
\vspace{-0.5em}
\renewcommand\arraystretch{1.2} 
\begin{threeparttable}
\begin{tabularx}{0.85\linewidth}{l|p{3.0mm}|p{15.5mm}|p{11.0mm}|p{7.0mm}}
\hline
Method & P. & Model & Strategy & DR \\
\hline
\cite{pan2020safe, cui2021lookout,chen2022interactive}  & $\checkmark$ & HDF; & CA; RA; & - \\
\hline
\cite{tian2020game, chandra2022gameplan} & $\times$ & Game theory; & CA; RA; & L-F \\
\hline
\cite{zhang2020efficient, luo2021interactive} & $\times$ & HDF; MDPs; & CA; RA; & - \\
\hline
\cite{crosato2022interaction, espinoza2022deep, rosbach2019driving, huang2023conditional} & $\times$ & Network & Implicit & - \\
\hline
\rowcolor[HTML]{E7E6E6}
\;Ours & $\checkmark$ & HDF; & CA; & I-R \\
\hline
\end{tabularx}
\smallskip
\scriptsize
\begin{tablenotes}
\RaggedRight
P. : the utilization of prediction results; DR: distinct roles in dealing with traffic agents; HDF: hand-designed functions; CA: collision avoidance; RA: risk avoidance; L-F: leader-follower model; I-R: influencer-reactor.\\
\end{tablenotes}
\end{threeparttable}

\renewcommand\arraystretch{1.0} 

\vspace{-1.5em}
\label{table:related_works}
\end{table}

\vspace{-0.5em}
\subsection{Learning-based Methods}

Learning-based methodologies have gained significant attention in the field of autonomous driving due to their effective understanding of complex environments. For example, learning-based behavioral planners\cite{crosato2022interaction, espinoza2022deep} had developed policies that take into account interactions, offering comprehensive guidance on cooperative behaviors of other vehicles. In contrast, methods based on inverse reinforcement learning (IRL) \cite{rosbach2019driving, huang2023conditional} aimed to derive underlying cost functions from expert demonstrations, allowing for the imitation of human-like interaction behaviors. 
In these learning methods, interactions between agents are typically modeled using relevance values in attention-based encoding techniques \cite{liang2020learning, gu2021densetnt}, where a larger value indicates a stronger interaction relation. While these methods have made significant advancements in autonomous driving by exploring various aspects of traffic interactions, they primarily focus on modeling interaction outcomes rather than the actual processes of interaction. As a result, they lack an explanatory framework for agents' actions from an interaction perspective. In contrast, our approach achieves a mathematical modeling of traffic interactions by explicitly defining interaction relations and formulating related conditions and constraints. Additionally, we conduct control experiments to verify the effectiveness of interaction-related coefficients, thereby enhancing the interpretability and configurability of the method for application in new scenarios.

Overall, as presented in Table \ref{table:related_works}, we conducted a comparative analysis between our method and other existing methods to highlight the advantages and distinctions of our approach.


\vspace{-0.5em} 
\section{System Overview \label{sect:sys_overview}}

\begin{figure}[!t]
    \centering
    \includegraphics[width=0.75\linewidth]{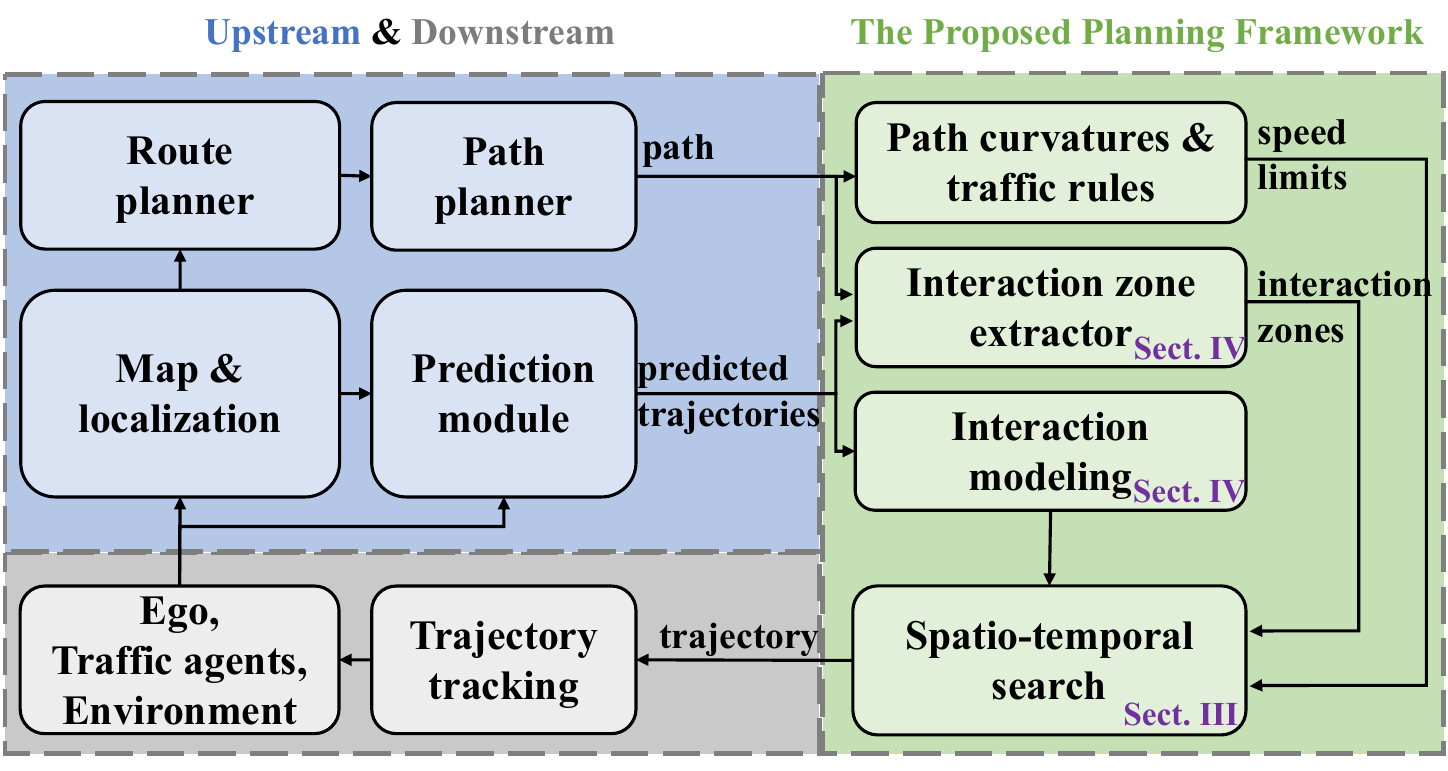}
    \vspace{-0.5em}
    \caption{Structures of the proposed planning framework (on the right) and its connections with other system components. }
    \label{fig:framework}
\vspace{-0.5em}
\end{figure}

As shown in Fig. \ref{fig:framework}, the entire system in this work employs a path-speed decoupled planning framework to calculate the trajectory for the AVs. The planning process involves generating the path while considering the road structure and subsequently planning the longitudinal motion using the proposed interactive s-t searcher. 
The rest of this paper is organized as follows: In Sect. \ref{sect:planning_pipeline}, we introduce the fundamental solutions to the planning problem, encompassing state representations, path generation, and spatio-temporal search. 
Sect. \ref{sect:interaction_reasoning} delves into the process of modeling traffic interactions within a spatio-temporal search framework. 
Sect. \ref{sect:simulation_env} provides the experimental simulation environment utilized in this study.	In Sect. \ref{sect:experiments}, we present comprehensive implementation details, including modules of localization and trajectory tracking. Moreover, this section encompasses the experimental results obtained from simulations, accompanied by corresponding discussions. Lastly, Sect. \ref{sect:discussion_conclusion} offers the concluding remarks for this paper.

\begin{table}[!t]
\centering
\caption{Nomenclatrue}
\vspace{-0.5em}
\renewcommand\arraystretch{1.0} 
\begin{tabularx}{\linewidth}{p{0.05\linewidth}| p{0.9\linewidth}}
\hline
$\bm{s}$ & longitudinal state vector of the vehicle along the given path. \\
$s$ & longitudinal position of the vehicle along the given path. \\
$v$ & longitudinal speed of the vehicle. \\
$a$ & longitudinal acceleration of the vehicle. \\
$j_{p,c}$ & estimated jerk value given the parent-child nodes $\bm{s}_p$ and $\bm{s}_c$. \\ 
\end{tabularx}
\renewcommand\arraystretch{1.2} 
\begin{tabularx}{\linewidth}{p{0.05\linewidth}| p{0.9\linewidth}}
$\bm{x}_{i,j}$ & the $j$-th prediction trajectory of the agent $i$. \\
$x_{i,j}^n$ & the $n$-th discrete state of the $\bm{x}_{i,j}$. \\
$t_{i,j}^n$ & time stamp value of $x_{i,j}^n$. \\
\hline
\end{tabularx}
\renewcommand\arraystretch{1.0} 

\renewcommand\arraystretch{1.0} 
\begin{tabularx}{\linewidth}{p{0.05\linewidth}| p{0.9\linewidth}}
$c_T$ & the search horizon of time in forward search. \\
$c_v$ & the minimum speed condition in forward search. \\
$c_s$ & the search horizon of accumulated distance. \\
$c_t$ & safety time boundary for collision avoidance. \\
$c_{z1}$ & the maximum distance interval for defining the interaction zone. \\
$c_{z2}$ & the maximum coverage distance of an inverse interaction zone. \\
$w_{(\cdot)}$ & coefficients of the cost function in search. \\
\end{tabularx}

\renewcommand\arraystretch{1.2} 
\begin{tabularx}{\linewidth}{p{0.05\linewidth}| p{0.9\linewidth}}
$[\underline{a}, \overline{a}]$ & longitudinal acceleration bounds of the AV. \\
$[\underline{j}, \overline{j}]$ & jerk bounds of the AV. \\
$\overline{a}_{\mathrm{lat}}$ & the maximum lateral acceleration of the AV. \\
$\underline{a}_i$ & the minimum acceleration of the agent $i$ when it reacts to the AV. \\
\hline
\end{tabularx}
\renewcommand\arraystretch{1.0} 
\label{table:notations}
\vspace{-1.5em}
\end{table}


\vspace{-0.5em}
\section{Problem Formulation \label{sect:planning_pipeline}} %

Several assumptions are made to enable the planning process for the AVs, including having prior knowledge of the road network and the reference route of the environment, as well as access to the position, speed, and acceleration of the AV, along with the shape, position, heading, and speeds of other traffic agents. 
In this study, the planning problem is addressed using a general pipeline planning framework \cite{lim2019hybrid}, which consists of two main steps. Firstly, a path is calculated to merge into a given reference route using the Fren\'et Frame. Secondly, a forward search method is applied along the generated path with consideration of multi-modal prediction results. The key notations utilized in this paper are listed in Table \ref{table:notations}.

\vspace{-0.5em}
\subsection{Path Generation in Fren\'et Frame \label{subsect:frenet representaion}}

\begin{figure}[!t]
    \centering
    \includegraphics[width=0.6\linewidth]{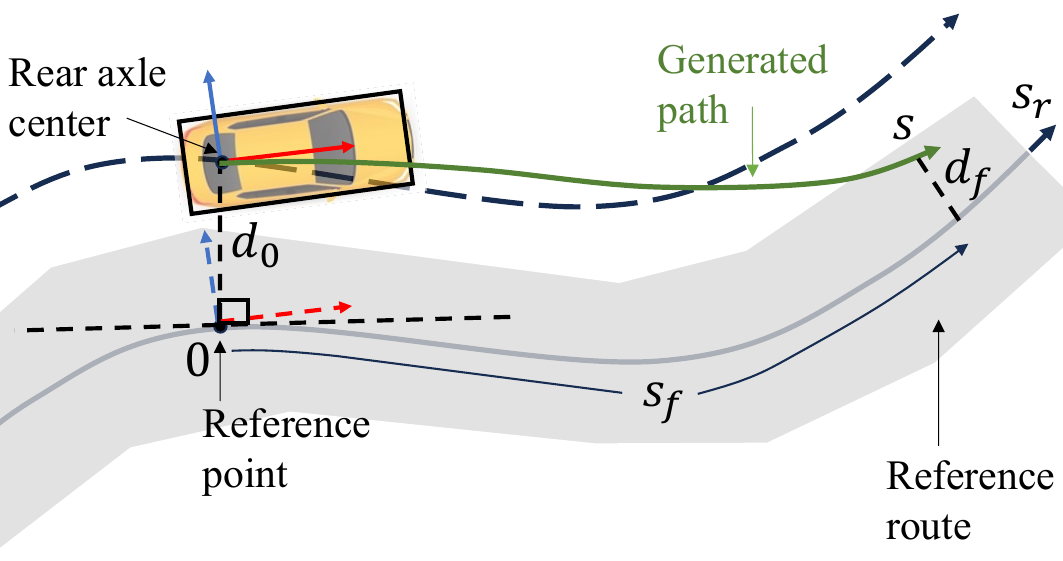}\vspace{-0.5em}
    \caption{An example of the AV and its path merging with the reference route.}\
    \label{fig:frenet_representation}
\vspace{-1.5em}
\end{figure}

In path generation, we model the state of the AV at its rear axle center by utilizing Frenet Frame \cite{werling2010optimal} representations along the reference route. As shown in Fig. \ref{fig:frenet_representation}, let $s_r$ denotes the AV's longitudinal accumulated distance along the reference route, $\bm{d}(s_r)=[d, d^{\prime}, d^{\prime\prime}]$ represents the lateral states, where $d^{\prime}$ and $d^{\prime\prime}$ are the first and the second derivatives. Then, the AV's path is formulated as the quintic polynomial \cite{lim2019hybrid, zhang2021hierarchical} 
\begin{equation}
    d(s_r) = c_0 + c_1 s_r + c_2 s_r^2 + c_3 s_r^3 + c_4 s_r^4 + c_5 s_r^5,
\end{equation}
\noindent where $c_{(\cdot)}$ are coefficients. In practice, we assume that the AV's initial $s_r$ is $0$, $\bm{d}(0)$ is $[d_0, d^\prime_0, d^{\prime\prime}_0]$, and the terminal $s_r$ is $s_f$, along with $\bm{d}(s_f) = [d_f, d^\prime_f, d^{\prime\prime}_f]$. Then, the coefficients are obtained by $[c_0, c_1, c_2] = [d_0, d^\prime_0, \frac{1}{2} d^{\prime\prime}_0]$ and 
\begin{align}
    \begin{bmatrix}
    c_3 \\
    c_4 \\
    c_5 \\
    \end{bmatrix} &= \begin{bmatrix}
    s_f^3 & s_f^4 & s_f^5\\
    3s_f^2 & 4s_f^3 & 5s_f^4\\
    6s_f & 12s_f^2 & 20s_f^3\\
    \end{bmatrix}^{-1} \begin{bmatrix}
    \Delta d - d^\prime_0 s_f - \frac{1}{2} d^{\prime\prime}_0 s_f^2 \\
    d^\prime_f - d^\prime_0 - d^{\prime\prime}_0 s_f \\
    d^{\prime\prime}_f - d^{\prime\prime}_0 \\
    \end{bmatrix},
\end{align}
\noindent where $\Delta d = d_f - d_0$. We set $\bm{d}(s_f) = \bm{0}$ to enable the path to merge into the reference route. In addition, multiple $s_f$ values are sampled to obtain a desired path, achieving a balance between smoothness and merging distance. 

\vspace{-0.5em}
\subsection{Spatio-temporal Planning\label{subsect:s-t planning}}

\begin{figure}[!t]
\centering
\vspace{-0.25em}
\includegraphics[width=0.8\linewidth]{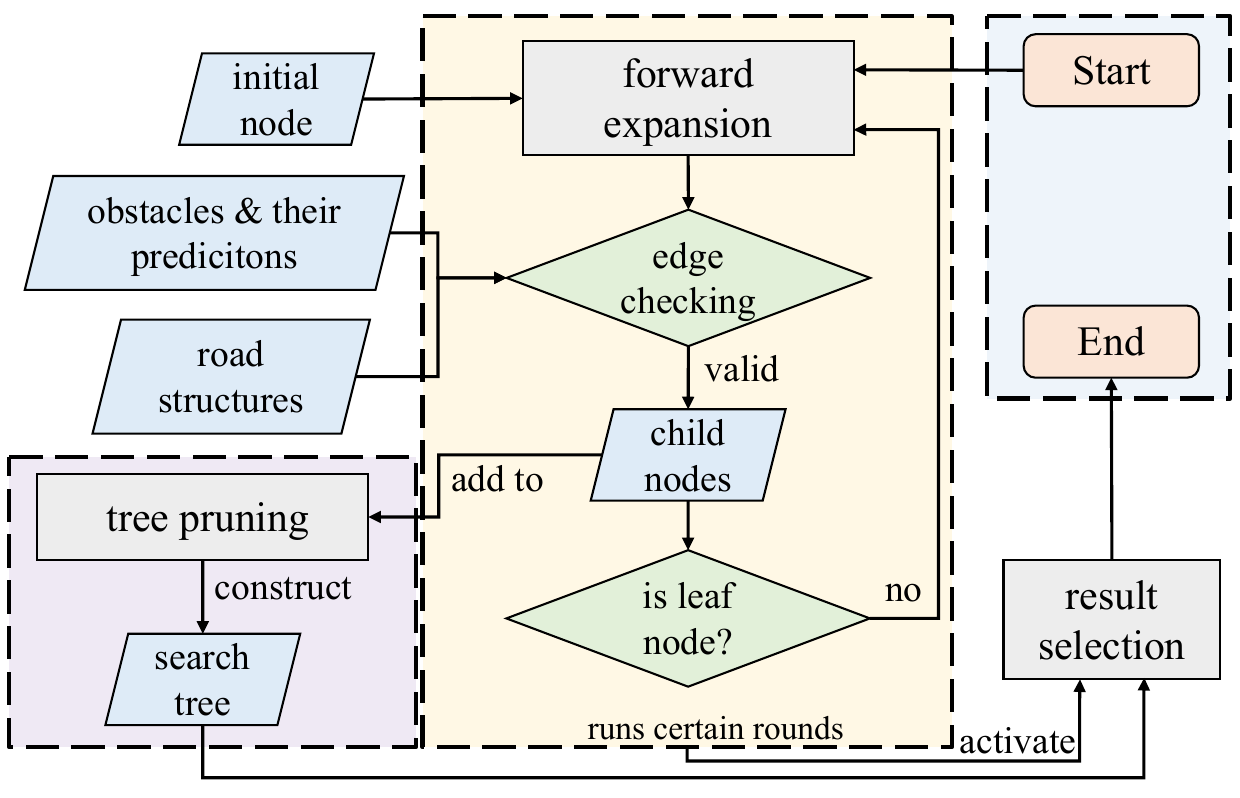}\vspace{-0.5em}
\caption{Pipeline of general s-t searcher for AVs.}
\label{fig:st-planner pipepline}
\vspace{-1.5em}
\end{figure}

To provide context for our proposed method, we first review the prevalent s-t forward search framework (depicted in Fig. \ref{fig:st-planner pipepline}). The s-t search is conducted to determine a valid speed profile within the Fren\'et frame of the path created in the preceding section, considering multi-modal prediction results.
Let $\bm{s}(t)=[t, s, v, a]$ represents the system state of the AV, where $t$ signifies the timestamp, $s$ indicates the longitudinal position, $v$ and $a$ are the speed and acceleration respectively. For a traffic agent, let $\bm{x}_{i,j} = \{x_{i,j}^n | n \in \{1, ..., N\} \} \in \mathbb{R}^{5\times N}$ denote the $j$-th prediction trajectory of the agent $i$, comprising of $N$ discrete states. Each state comprises 5-dimensional data, encompassing the accumulated distance $s_{i,j}^n$ (along the predicted trajectory), timestamp $t_{i,j}^n$, and 3-D positions of the agent. Several constraints are involved during the search, including the AV's acceleration bounds $[\underline{a}, \overline{a}]$, speed limit $\overline{v}$ dictated by traffic rule, and the curvature speed constraint
\begin{equation}
    \overline{v}_{\kappa}(s) \leq \sqrt{\overline{a}_{\mathrm{lat}} / |\kappa(s)| },
\end{equation}
\noindent where $\overline{a}_{\mathrm{lat}}$ denotes the lateral acceleration limit, and $\kappa(s)$ indicates the path curvature. Furthermore, longitudinal jerk constraints is defined as $[\underline{j}, \overline{j}]$.
During the forward search process, a tree structure is employed to represent the connections between visited states. Each node in the tree corresponds to a discrete state $\bm{s}(t)$ of the AV. The forward search involves four components: forward expansion, edge checking, tree pruning, and cost function to facilitate the tree search.

\subsubsection{Forward Expansion}

We expand the child nodes using the discrete spatial formulation \cite{li2021spatial,chen2022efficient}, which allows the proposed method to process spatial interaction information together (details see Sect. \ref{subsect:interaction_edge-check}). Given a parent node $\bm{s}_p$, a child node $\bm{s}_c$ is obtained by constant acceleration model (CAM)
\begin{align*}
    &\begin{bmatrix}
     t_c \\ s_c \\ v_c \\ a_c
    \end{bmatrix} = \begin{bmatrix}
        t_p + 2 \Delta s / (v_c + v_p) \\
        s_p + \Delta s \\
        (v_p^2 + 2 u \Delta s)^\frac{1}{2} \\
        u
    \end{bmatrix}, \\ 
    \text{s.t. } &v_c \leq \min(\overline{v}, \frac{1}{\Delta s}\int_{s_c}^{s_p} \overline{v}_{\kappa}(s) ds ), \\
    &u \in {U} \subset [\max(\underline{a}, v_p^2 / (2 \Delta s)), \overline{a}], \\
    &j_{p,c} = (a_c - a_p)/(t_c - t_p) \in [\underline{j}, \overline{j}], \tag{\theequation} \label{equation:vehicle_model} \refstepcounter{equation}
\end{align*}
\noindent where $\Delta s$ represents the forward sampling distance along the target path, estimated using the heuristic function from \cite{chen2022efficient}. The acceleration $u$ is selected from a set of discretized control inputs ${U}$, and $j_{p,c}$ is the estimated jerk to control the smoothness of the output trajectory. 
Moreover, a spawned node will be treated as a leaf node that not allowed be expanded forward if its timestamp exceeds a designated planning horizon (denoted as $c_T$), if the speed falls below a specified threshold $c_v$, or if the distance horizon exceeds a constant value $c_s$.

\subsubsection{Edge Checking \label{subsubsect:edge_checking}}

Given a parent node $\bm{s}_p$ and its expanded child node $\bm{s}_c$, collision-avoidance based methods promise the generated node and its edge obeying
\begin{align*}
    | t_{i,j}^n - t_k| \geq c_t, 
    \forall x_{i,j}^n \in X_k, \forall s_k \in {S}_{p,c} \subset [s_p, s_c], \tag{\theequation} \label{equation:collision_constraint} \refstepcounter{equation}
\end{align*}
\noindent where ${S}_{p,c}$ is a discrete sample set $\{s_1, ..., s_K\}$ within the interval $[s_p, s_c]$, $s_k$ is the $k$-th sample value, and $X_k$ denotes the set of predicted states that overlap with the AV at $s_k$ (considering shape factors). In addition, $t_k$ is the interpolated timestamp value of the checked edge at $s_k$, and $c_t$ is a constant used to define the safety time boundary for collision avoidance.

\subsubsection{Tree Pruning}

Because the number of nodes increases exponentially during the process of forward expansion, we employ the local truncation technology used in our previous work \cite{chen2022efficient}. In this approach, we prune all branches that have terminals at the same $(s, t, v)$ grids, retaining only the best one. In the subsequent section, we will introduce the cost function  used to determine the best node. 

\subsubsection{Cost Function \label{subsubsect:cost_func}}

For a spawned child node $\bm{s}_c$ from its parent node $\bm{s}_p$, its cost is defined as
\begin{equation}
J_c = J_p + w_v J_v + w_a J_a + w_j J_j, 
\end{equation}
\noindent where $J_p$ is the cost of the parent node and $w_{(\cdot)}$ denotes the coefficients for each cost item. In addition, $J_v = (|\overline{v} - v_c| \cdot \Delta t)$ denotes the cost of the deviation from the speed limit, where $\Delta t = (t_c - t_p)$ is used as a scale factor. $J_a = (a_c^2 \cdot \Delta t)$ indicates the control efforts of the AV, and $J_j = (j_{p,c}^2 \cdot \Delta t)$ influences the smooth of the trajectory.


\vspace{-0.2em} 
\section{Interactive S-t Planning\label{sect:interaction_reasoning}}

We embed interaction modeling in the traditional planning framework outlined in Sect. \ref{subsect:s-t planning}. Specifically, we model traffic interactions at each searched tree node by carefully considering their interaction relations with predicted trajectories.

\vspace{-0.5em}
\subsection{Definitions of Interaction Relations  \label{subsect:interaction_def}}

Collision-based methods implicitly model interactions between agents using two relational descriptions: overtaking and yielding. These descriptions indicate which agent passes first when they encounter the same conflict spaces. Inspired by Sun's work \cite{sun2022m2i}, we introduce the concepts of influencer and reactor to model relations of interacting agents in our study. However, our approach differs in one \textbf{major aspect}: In our method, the interaction relation is modeled at the trajectory state level, rather than at the trajectory or agent level. This design permits the flexibility for a trajectory to operate as both a reactor and an influencer in different stages.


We categorize interactions into two main classes: \textit{influencer} and \textit{reactor}. The \textit{influencer} relation describes scenarios where the AV's motion can influence or change the motion of other agents. The \textit{reactor} relation pertains to situations where the AV needs to react to the motions of other agents, which further includes two sub-classes: \textit{overtaking} and \textit{yielding}.

\textit{Definition 1}: We refer to the relation as \textit{overtaking} when the planned state $\bm{s}_k = [t_k, s_k, v_k, a_k]$ reacts to the predicted state $x_{i,j}^n$ but arrives before it, which can be formulated by
\begin{equation}
t_k - t_{i,j}^n \leq -c_t.
\label{equation:overtake_constraint}
\end{equation}

\textit{Definition 2}: For the \textit{yielding} relation, the planned state $\bm{s}_k$ arrives at $x_{i,j}^n$ after agent $i$, and the formulation becomes 
\begin{equation}
t_k - t_{i,j}^n \geq c_t.
\label{equation:yiled_constraint}
\end{equation}
\noindent The combination of (\ref{equation:overtake_constraint}) and (\ref{equation:yiled_constraint}) is (\ref{equation:collision_constraint}), indicating that the ``reactor'' relation employs a collision-avoidance strategy.

\textit{Definition 3}: To describe the situation when the planned state $\bm{s}_k$ influences $x_{i,j}^n$, we define the \textit{influencing} constraint
\begin{equation}
    |t_{i,j}^n(u_{i,j}) - t_k| \geq c_t, \exists u_{i,j} \in [\underline{a}_i, \overline{a}_i]. 
\label{equation:influ_constraint}
\end{equation}
\noindent The response of the influenced agent $i$ is is represented by its longitudinal acceleration $u_{i,j}$ along the predicted trajectory. In this model, $t_k$ is considered invariant under the assumption that agent $i$ is aware of the AV's expected arrival time at $s_k$. In addition, the response $t_{i,j}^n$ is obtained using CAM as
\begin{align}
    t_{i,j}^n(u_{i,j}) &= \frac{-v_{i,j}^0 + ((v_{i,j}^0)^2 + 2 u_{i,j} \cdot s_{i,j}^n)^{\frac{1}{2}}}{2 u_{i,j}}, \notag \\
    \text{s.t. } u_{i,j} &\in [\max (\underline{a}_i, -(v_{i,j}^0)^2 / (2 s_{i,j}^n)), \overline{a}_i],
\end{align}
\noindent where $v_{i,j}^0$ denotes the initial speed of the agent $i$, $[\underline{a}_i, \overline{a}_i]$ represents the response acceleration bounds for the agent $i$ when influenced by the AV, and $u_{i,j}$ is the average acceleration along $\bm{x}_{i,j}$. 
As modeled in (\ref{equation:influ_constraint}), when an agent's movement is influenced by the AV, two options may arise: speeding up or slowing down. 
In this work, our sole focus is on scenarios in which the AV induces reactive deceleration motions in other interacting agents. As a result, (\ref{equation:influ_constraint}) is simplified as
\begin{equation}
    (t_{i,j}^n(u_{i,j}) - t_k) \geq c_t, \exists u_{i,j} \geq \underline{a}_i.
\label{equation:influ_constraint_v2}
\end{equation}
\subsection{Interaction Zone \label{subsect:interaction_zone}}

\begin{figure}[tb]
    \centering
    \subfloat[Point overlap]{
        \includegraphics[width=0.4\linewidth]{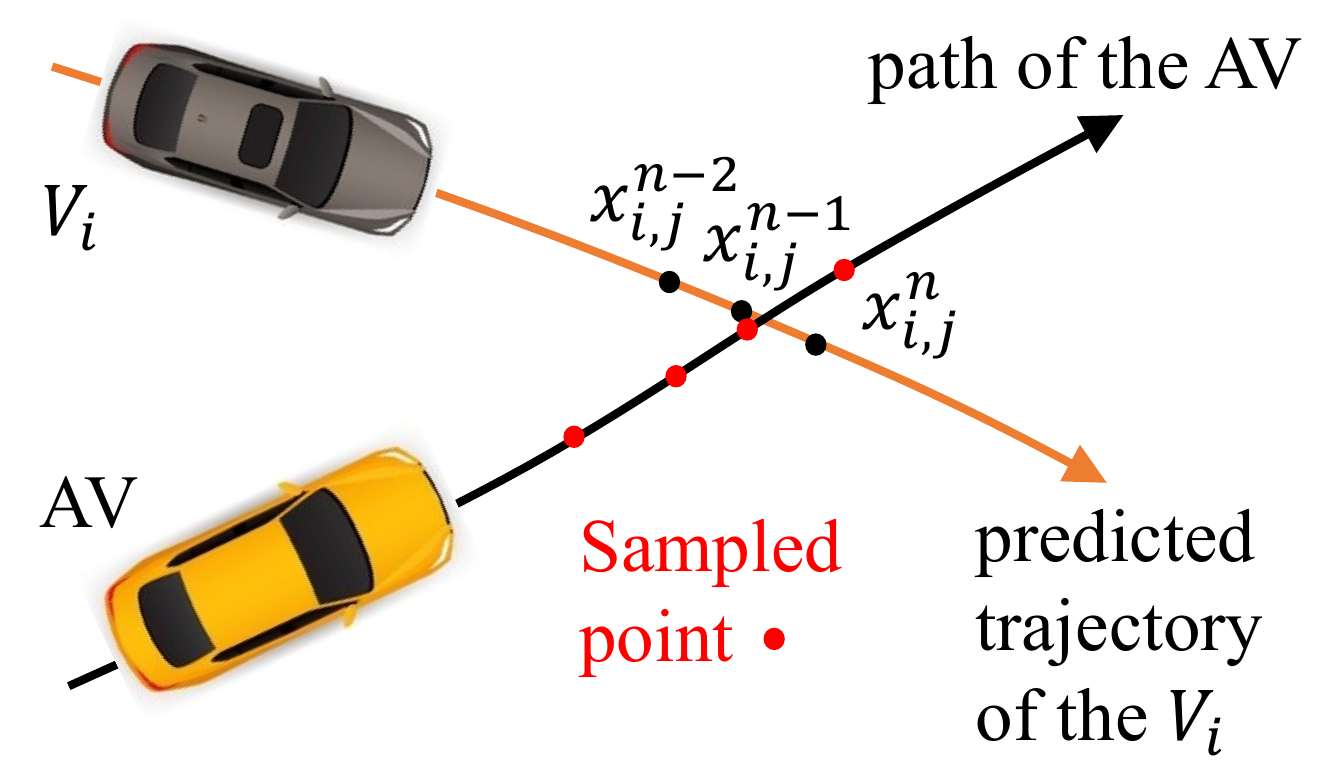}
        \label{subfig:point_overlap}} \hspace{1.5em}
    \subfloat[Line overlap]{
        \includegraphics[width=0.4\linewidth]{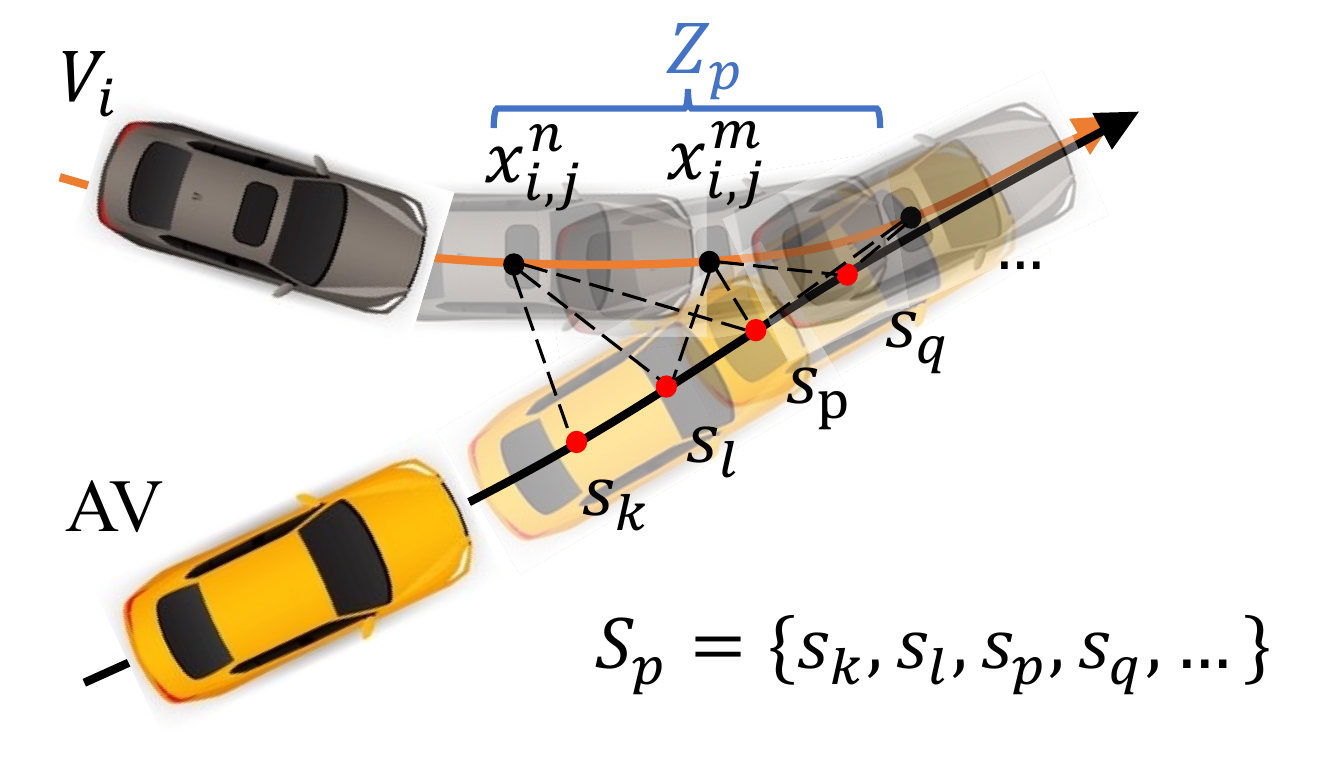}
     \label{subfig:line_overlap}} \\
    \subfloat[Oncoming encounter]{
        \includegraphics[width=0.4\linewidth]{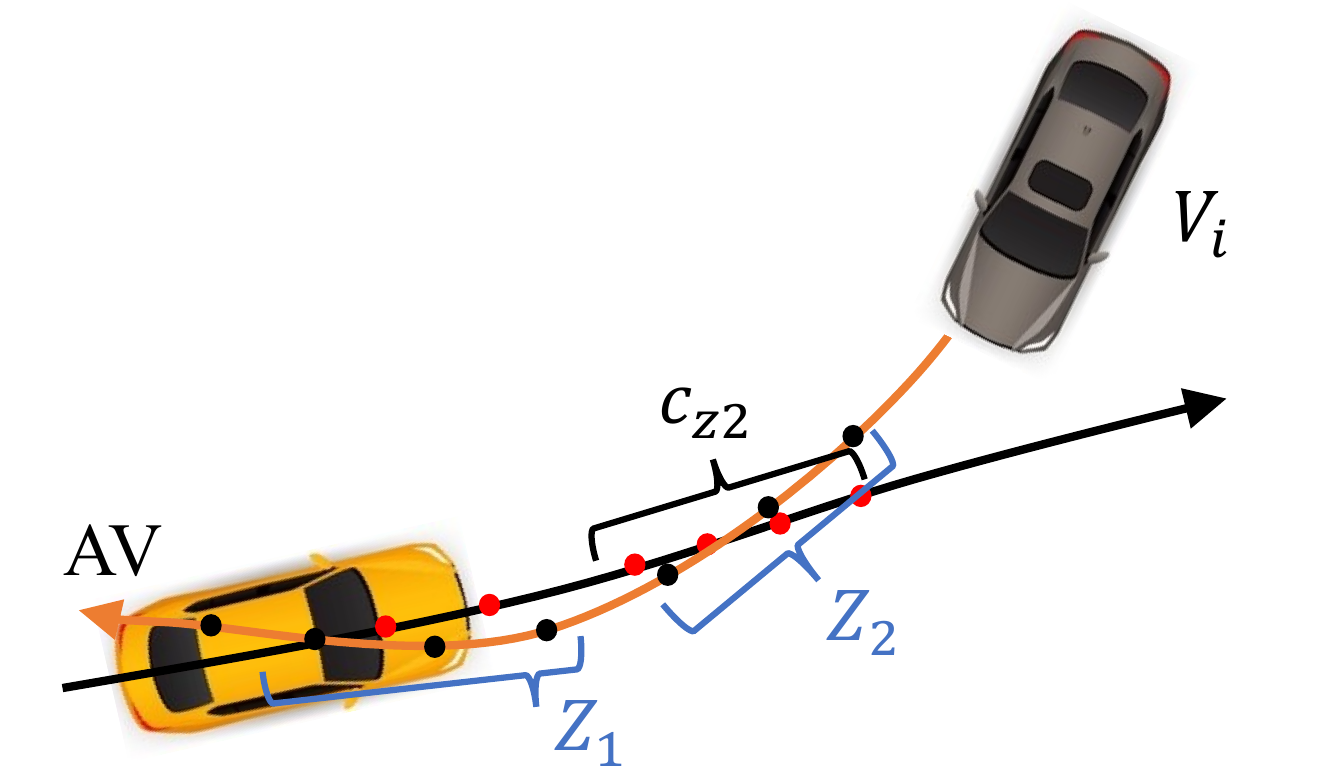}
        \label{subfig:inv_overlap}} \hspace{1.5em}
    \subfloat[Special situation]{
        \includegraphics[width=0.4\linewidth]{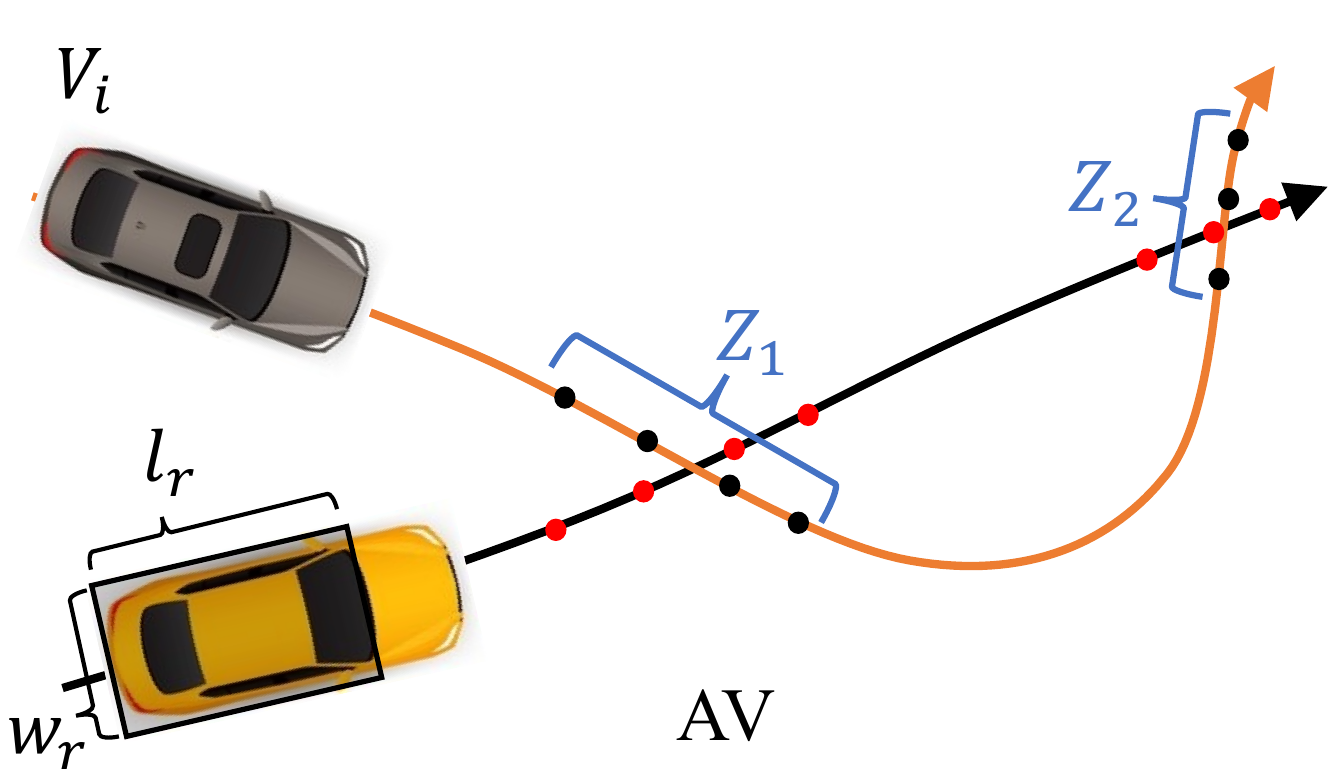}
        \label{subfig:overlap_special_situ}}
    \caption{Illustrations of interactions and interaction zones ${Z}_1$ and ${Z}_2$.}
    \label{fig:interaction_situations}
\vspace{-1.5em}
\end{figure}

In \ref{subsect:interaction_def}, we introduce diverse interaction relations for a planned state $\bm{s}_k$ and predicted state ${x_{i,j}^n}$. However, utilizing the predicted state as the subject for modeling interactions is generally inappropriate, as the number of states in a predicted trajectory is typically greater than its corresponding interaction relations.
Fig. \ref{fig:interaction_situations} illustrates several examples. In Fig. \ref{subfig:inv_overlap}, the $Z_1$ part of predicted states shares a common interaction relation, indicating that the agent should yield to the AV's trajectory within that part. Meanwhile, the interaction relations of the $Z_2$ segment need further determination. Another example is shown in Fig. \ref{subfig:overlap_special_situ}, where the relations of nonadjacent states in the $Z_1$ and $Z_2$ parts may differ.
As a result, we introduce the concept of the \textit{interaction zone} to categorize a set of prediction states that share the same interaction relation.

\textit{Definition 4}: For a predicted result $\bm{x}_{i,j}$, we group its states $\{x_{i,j}^n : \bm{x}_{i,j}\}$ into an \textit{interaction zone} $Z_p$ if they satisfy
\begin{equation}
\exists |s_k - s_l| \leq c_{z1}, \forall s_k, s_l \in S_p,
\label{equation:iarea_def}
\end{equation}
\noindent where $S_p$ represents the set of the AV's states that overlap with predicted states ${x_{i,j}^n \in Z_p }$ (as depicted in Fig. \ref{subfig:line_overlap}), and $c_{z1}$ is a constant indicating the maximum distance interval for being considered inside the same interaction zone. However, (\ref{equation:iarea_def}) may not be applicable in an oncoming encounter situation, as demonstrated in Fig. \ref{subfig:inv_overlap}. To address this, we introduce the concept of an \textit{inverse interaction zone}, which occurs when at least one state has a heading opposite to the AV's direction. In the inverse interaction zone, it further adheres to 
\begin{equation}
    |s_k - s_l| \leq c_{z2}, \forall s_k, s_l \in S_p, \\
\end{equation}
\noindent where $c_{z2}$ is a constant representing the maximum coverage distance of an inverse interaction zone along the AV's path.

\vspace{-0.5em}
\subsection{Relation Determination \label{subsect:relation_deter}}


After defining the interaction relation and the interaction zone, the next step is to judge the relations for each interaction zone. In this study, the relations are established in two stages: firstly, before the planning process, and secondly, during the planning phase. Once a relation is determined for an interaction zone, it remains unchanged in subsequent calculations.

\subsubsection{Relation judgement before planning \label{subsubsect:relation_deter_before_plan}}
In real-world scenarios, certain relations are initially established. For example, when the AV follows an agent, it is originally defined as the AV yielding to that agent. Conversely, when an agent follows the AV, it is the AV influencing the agent. Following symbols from (\ref{equation:overtake_constraint}) and (\ref{equation:iarea_def}), we generalize these situations using 
\begin{equation}
r(Z_p) = \begin{cases}
R_f, & \text{if } s_k \leq 0, t_{i,j}^n \geq c_t, \exists x_{i,j}^n \in Z_p,  \\
R_y, & \text{if } s_k > 0, t_{i,j}^n < c_t, \exists x_{i,j}^n \in Z_p, \\
R_u, & \text{otherwise}, 
\end{cases}
\label{equation:relation_init_deter}
\end{equation}
\noindent where $r(Z_p)$ represents the interaction relation of zone $Z_p$. Specifically, $s_k \leq 0$ indicates the AV overlaps with predicted states $x_{i,j}^n \in Z_p$ at its initial location, and $s_k > 0$ and $t_{i,j}^n < c_t$ indicate that the agent initially occupies the AV's path at $s_k$. In this step, the shape intersection calculations solely consider the rear part of the AV, with a reduced width of $w_r$ and length of $l_r$ (as shown in Fig. \ref{subfig:overlap_special_situ}). Moreover, the symbols $R_{(\cdot)}$ are the interaction relations, where $R_f$ represents the \textit{influence} relation, $R_y$ denotes the yielding relation, and $R_u$ means that the relation of $Z_p$ is undetermined. 

\subsubsection{Relation judgement during planning \label{subsubsect:rela_judge_dur_plan}} 
In the planning process, when determining the undetermined relation $r(\bm{s}_k, x_{i,j}^n)$ of a specific tree node $\bm{s}_k$, common collision avoidance methods implicitly establish the relation using the collision constraint defined in (\ref{equation:collision_constraint}), which can be rephrased as
\begin{equation}
r(\bm{s}_k, x_{i,j}^n) = \begin{cases}
R_o, & \text{if } t_k \leq t_{i,j}^n - c_t,  \\
R_y, & \text{if } t_k \geq t_{i,j}^n + c_t, \\
R_{\times}, & \text{otherwise,}
\end{cases}
\label{equation:pred_judge_relation}
\end{equation}
\noindent where $R_o$ represents the overtaking relation, and $R_{\times}$ indicates that the corresponding relation is invalid. 

As (\ref{equation:influ_constraint_v2}) assumes that the influenced agent can immediately respond to the AV's motion with an acceleration $u_{i,j}$, which is often unrealistic. Hence, we further define condition
\begin{equation}
    t_k + c_{f1} + \frac{c_{f2}}{v_k} \leq t_{i,j}^n,
\label{equation:influ_constraint2} 
\end{equation}
\noindent where $c_{f1}$ and $c_{f2}$ are constant coefficients, $c_{f1}$ term represents a time gap coefficient, and $c_{f2} / v_k$ term encourages the AV to fast pass the interaction zone to establish an \textit{influence} relation. As a result, our method judges the relation by 
\begin{equation}
r(\bm{s}_k, x_{i,j}^n) = \begin{cases}
R_f & \text{if both (\ref{equation:influ_constraint_v2}) and (\ref{equation:influ_constraint2}) are met}  ,\\
R_o, & \text{elif } t_k \leq t_{i,j}^n - c_t,  \\
R_y, & \text{elif } t_k \geq t_{i,j}^n + c_t, \\
R_{\times}, & \text{otherwise.}
\end{cases}
\label{equation:irule_judge_relation}
\end{equation}
\noindent It is important to note that the parameter $\underline{a}_i$ used in (\ref{equation:influ_constraint_v2}) and (\ref{equation:irule_judge_relation}) has different meanings. They indicate how the AV reacts to agents' potential reactions before and after an influence relation is established, respectively. In addition,  (\ref{equation:irule_judge_relation}) embodies a fundamental design principle in our methodology, which entails retaining the prediction results of other agents when yielding to them and modifying the prediction outcomes based on their longitudinal reactions when exerting influence.

Above (\ref{equation:pred_judge_relation}) and (\ref{equation:irule_judge_relation}) calculate the interaction relation for each predicted state, but not for their corresponding interaction zones. Since one planned state might intersect with multiple predicted states within an interaction zone, a group of predicted states $\{x_{i,j}^n: Z_p \}$ may be collectively evaluated to establish the relation for an interaction zone $Z_p$. This can result in diverse relation outcomes for a single interaction zone. We will address the appropriate solutions in Sect. \ref{subsect:interaction_edge-check}, where multiple nodes from the same tree edge are collectively processed to update the interaction relations of a child node.

\subsection{Node Representation \label{subsect:interaction_node-def}}

In the conventional framework discussed in Section \ref{subsect:s-t planning}, a tree node encompasses state information $\bm{s}(t) = [t, s, v, a]$ of the AV, along with supplementary auxiliary data such as the discrete node index and the index of its parent node, aiding the search process. In contrast, in our approach, each node additionally stores its relationships with existing interaction zones. As mentioned in Sect. \ref{subsubsect:relation_deter_before_plan}, these relationships are initially established before the planning commences. Subsequently, during planning, these relationships in child nodes are inherited from the parent node and may be modified during the planning process if they remain undetermined.

\subsection{Edge Checking \label{subsect:interaction_edge-check}}

\begin{figure}[!t]
\vspace{-0.7em}
\begin{algorithm}[H]
\caption{Edge Checking with Interaction Relations}\label{alg:interaction_edge_checking}
\begin{algorithmic}[1]
    \State \textbf{Notation}: parent node's relations $\bm{r}_p$, interaction zone amount $P$, predicted states $\mathbf{Y}_K$ and their corresponding planned nodes $\{\bm{s}_k : \bm{S}_K\}$ at $\{s_1, s_2, ..., s_K\}$, nodes relations matrix $\mathbf{R}_{K}$, and validity matrix $\mathbf{V}_{K}$.
    \State \textbf{Initialize}: $\bm{S}_K \gets$ \textsc{Interp}($\bm{s}_p$, $\bm{s}_c$), $\mathbf{R}_{K} \gets \bm{r}_p \times K$.

    \State $\mathbf{L}_u \gets$ \textsc{EqualTo}($\mathbf{R}_{K}$, $R_u$)
    \State \textsc{Update}($\mathbf{R}_{K}[\mathbf{L}_u], \bm{S}_K, \mathbf{Y}_K$) using (\ref{equation:irule_judge_relation})
    \Statex \Comment{End relation determination}

    \State $\mathbf{V}_{K} \gets \mathbf{R}_{K} \neq R_{\times}$ \Comment{Initialize relation validities}
    \State $\mathbf{L}_f \gets$ \textsc{EqualTo}($\mathbf{R}_{K}$, $R_f$)
    \State \textsc{Update}($\mathbf{V}_{K}[\mathbf{L}_f], \bm{S}_K, \mathbf{Y}_K$) using (\ref{equation:influ_constraint_v2})
    \State $\mathbf{L}_o \gets$ \textsc{EqualTo}($\mathbf{R}_{K}$, $R_o$)
    \State \textsc{Update}($\mathbf{V}_{K}[\mathbf{L}_o], \bm{S}_K, \mathbf{Y}_K$) using (\ref{equation:overtake_constraint})
    \State $\mathbf{L}_y \gets$ \textsc{EqualTo}($\mathbf{R}_{K}$, $R_y$)
    \State \textsc{Update}($\mathbf{V}_{K}[\mathbf{L}_y], \bm{S}_K, \mathbf{Y}_K$) using (\ref{equation:yiled_constraint})
    \Statex \Comment{End relation constraint checking}

    \For{$p \in \{1, ..., P\}$}
        \State $R = \emptyset$
        \For{$k \in \{1, ..., K\}$ \textbf{if} \textsc{IsOverlapped}($\bm{s}_k$, $Z_p$)}
            \State $R = R \cup \mathbf{R}_{k}$
        \EndFor
        \State \textsc{Update}($\mathbf{V}_K$, $R$) using (\ref{equation:relation_conflict_solutions})
    \EndFor
    \Statex \Comment{End solving relation conflicts}
    \State is\_valid $\gets$ \textsc{Sum}($\mathbf{V}_K$) $\geq K$
    \State \textbf{return} is\_valid, $\mathbf{R}_K$
\end{algorithmic}
\end{algorithm}
\vspace{-2.0em}
\end{figure}

Different from the edge checking process in traditional s-t search, the proposed method performs three steps in this process, including relation determination mentioned in Sect. \ref{subsect:relation_deter}, relation constraint checking in Sect. \ref{subsect:interaction_def}, and resolution of relation conflicts and relation update for child nodes.

The edge checking process is shown in Algo. \ref{alg:interaction_edge_checking}. When a child node $\bm{s}_c$ is expanded from its parent node $\bm{s}_p$, we assume that $M$ predicted states with $P$ interaction zones are involved in this edge checking. Since one interaction zone may contain multiple predicted states ($M \geq P$), and one predicted state can have multiple intersections with the AV's path, we further assume there are $K$ ($K \geq M$) intersected planned states in the edge. As a result, we obtain $K$ interpolated tree nodes at locations $\{s_1, s_2, ..., s_K\}$, and $M$ predicted states are extended to $K$ predicted states $\mathbf{Y}_K$ for ease of calculation. This ensures a one-to-one interaction correspondence between elements in $\bm{S}_K$ and $\mathbf{Y}_K$, and their interaction relations, denoted as $\mathbf{R}_K$, are inherited from the parent node (Algo. \ref{alg:interaction_edge_checking}, Line 2).


Then, the undetermined elements of matrix $\mathbf{R}_K$ are updated using relation determination formulations (Lines 3-4), and a validity matrix $\mathbf{V}_K$ is initialized with ``true'' for locations where the corresponding relation is not invalid (Line 5). Subsequently, different constraint formulations are implemented to update the validity of the interpolated nodes. These constraints include influencing constraint (Lines 6-7), overtaking constraint (Lines 8-9), and yielding constraint (Lines 10-11) at their respective locations specified by $\mathbf{L}_{(\cdot)}$.
Finally, interaction relations $\mathbf{R}_K$ are collected together (Lines 12-18) to check for any conflicting interactions within an interaction zone $Z_p$, where Line 14 traverses the subset of indexes $\{1, 2, ..., K\}$ for which the corresponding tree nodes overlap with $Z_p$. In the example scenario in Fig. \ref{subfig:line_overlap}, if $\{x_{i,j}^n, x_{i,j}^m\}$ constitute the interaction zone $Z_{p}$, then the traversal involves $\{k, l, p, l, p, q\}$. Specifically, $x_{i,j}^n$ overlaps with planned nodes at $\{s_k, s_l, s_p\}$, while $x_{i,j}^m$ intersects at $\{s_l, s_p, s_q\}$. For relation conflicts (Line 17), mentioned in Sect. \ref{subsubsect:rela_judge_dur_plan}, they are resolved by
\begin{equation}
\begin{cases}
\mathbf{V}_K \gets \text{false} \times K, 
& \text{if $\{R_o, R_y\} \subset R$}, \\
\text{pass}, & \text{otherwise,}
\end{cases}
\label{equation:relation_conflict_solutions}
\end{equation}
\noindent where the existence of both influencing and yielding relations within the same interaction zone is not allowed. Finally, the child node is considered valid only if all edge nodes $\bm{S}_K$ are valid (Line 19). In practice, thanks to (\ref{equation:vehicle_model}), the entire process can be accelerated by parallelizing the processing of multiple expanded child nodes from a parent node.

For the returned relation matrix $\mathbf{R}_N$, its values will be used to revise the relation records of the $P$ interaction zones within the newly generated child node $\bm{s}_c$. Conflicts may emerge again when distinct relations are assigned to a single interaction zone. In such instances, we prioritize $R_f$ over $R_o$, and if both $R_f$ and $R_y$ are present, we establish the relation as $R_y$.

\vspace{-0.5em}
\subsection{Algorithm Complexity \label{subsect:iplanner-complexity}}


The update of relations in Algo. \ref{alg:interaction_edge_checking} introduces additional computational time to the planning process. To analyze the algorithm's complexity resulting from this operation, let's make the following assumptions: on average, $W$ parent nodes are expanded; for each parent node, an average of $N$ child nodes are generated; and an average of $K$ nodes are interpolated along the edge for validity checking.
Therefore, for the original collision checking, the computational complexity without any acceleration technologies would be $O(W \times N \times K)$ (Line 19 in Algo. \ref{alg:interaction_edge_checking}).
When considering interaction relations in planning, let's assume there are $N_R$ relations (excluding $R_\times$) involved. The complexity of validity checking in this case would be $O((N_R + 1) \times W \times N \times K)$. This includes $N_R$ times of determining relation locations in a matrix (Lines 6, 8, 10) and checking their constraints based on different types of relations (Lines 7, 9, 11). Once conflict resolution is involved (Lines 12-18), the computational complexity is approximately less than $O((N_R + 1) \times W \times N \times K + (P+1)\times K)$. Relevant experiments will be shown in Sect. \ref{subsubsect:exp_time_complexity}.


\section{Simulation}\label{sect:simulation_env}

We validated the effectiveness of our algorithms by implementing planning problems in the closed-loop simulation \cite{klischat2019coupling} within the commonroad environment \cite{althoff2017commonroad}. In the expriments, intersections are unprotected by the traffic lights to more effectively capture the impacts of prediction results on the planning performance. We simulated a total of 232 scenarios (80 seconds for each) by extracting them from the \textit{hand-crafted}, \textit{scenario-factory}, and \textit{SUMO} folders \cite{klischat2019coupling}. To ensure diversity, we restricted the selection to a maximum of 4 scenarios from each city. Moreover, we included 76 highly interactive intersection scenarios to facilitate comparisons of algorithms, considering the time-consuming and ineffective nature of implementing all simulations of the 232 scenarios.


\section{Experiments and Results}\label{sect:experiments}

\subsection{Implementation Details \label{subsect:imple_details}}

\begin{table}[!t]
\centering
\caption{Parameter values}
\vspace{-0.5em}

\renewcommand\arraystretch{1.0} 
\begin{tabularx}{\linewidth}{l|ccccccccc}
\hline
Params. & $c_T$ & $c_v$ & $c_s$ & $c_t$ & $c_{z1}$ & $c_{z2}$ & $w_v$ & $w_a$ & $w_j$ \\
\hline
Values & 6.0 & 0.1 & 100.0 & 0.5 & 5.0 & 5.0 & 5.0 & 0.5 & 0.8 \\
\hline
\end{tabularx}

\renewcommand\arraystretch{1.0} 

\vspace{-1.0em}
\label{table:common_parameter_values}
\end{table}

\begin{table}[!t]
\centering
\caption{Prediction validation results with 10 modes}
\vspace{-0.5em}

\renewcommand\arraystretch{1.0} 
\begin{threeparttable}
\begin{tabularx}{1.0\linewidth}{l|p{4.0mm}p{4.0mm}p{4.0mm}p{4.0mm}p{4.0mm}p{4.0mm}p{4.0mm}p{4.0mm}}
    \hline
    Pct. & 5\% & 10\% & 20\% & 30\% & 40\% & 60\% & 80\% & 100\% \\
    \hline
    $\text{MinADE}_{\text{4.0s}}$ & 0.94 & 0.80 & 0.71 & 0.69 & 0.66 & 0.61 & 0.61 & \textbf{0.58} \\
    \hline
    $\text{MissRate}_{\text{4.0s}}$ & 0.26 & 0.22 & 0.18 & 0.18 & 0.17 & 0.16 & 0.16 & \textbf{0.15} \\
    \hline
    $\text{MinADE}_{\text{6.0s}}$ & 1.83 & 1.56 & 1.42 & 1.34 & 1.28 & 1.20 & 1.21 & \textbf{1.17} \\
    \hline
    $\text{MissRate}_{\text{6.0s}}$ & 0.65 & 0.58 & 0.53 & 0.52 & 0.53 & 0.48 & 0.47 & \textbf{0.45} \\ 
    \hline
\end{tabularx}
\begin{tablenotes}
\RaggedRight
metrics are presented at two different horizons (4.0s, 6.0s) to better reflect the prediction performances.
\end{tablenotes}
\end{threeparttable}

\renewcommand\arraystretch{1.0} 

\vspace{-0.5em}
\label{table:pred_results}
\end{table}

\subsubsection{Planning}
\begin{figure}[!t]
    \vspace{0.15em}
    \centering
    \includegraphics[width=0.6\linewidth]{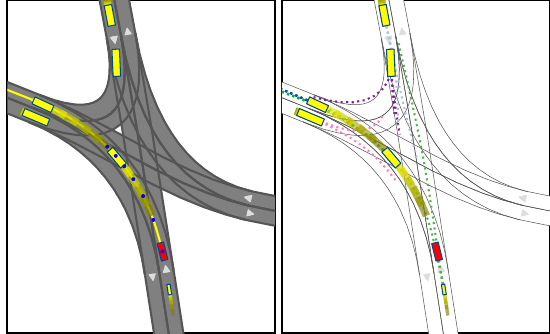}\vspace{-0.5em}
    \caption{An example simulation scenario with the AV represented by a red block, traffic agents as yellow blocks, prediction results as dashed lines on the right, and the AV's task route shown as a yellow solid line on the left.}
    \label{fig:simu_example_demos}
\vspace{-1.5em}
\end{figure}

The entire planning framework is developed using a Python backend. Specific parameters utilized in the experiments are detailed in Table \ref{table:common_parameter_values}.  The acceleration limit $[\underline{a}, \overline{a}]$ for the AV is $[-4.0, 3.0]$, the jerk bound $[\underline{j}, \overline{j}]$ is $[-8.0, 8.0]$, and the lateral acceleration limit $\overline{a}_{\mathrm{lat}}$ is $3.43\, m/s^2$. 

In the simulation, as shown in Fig. \ref{fig:framework}, the AV acquires the actual environment map and localization information from the simulation. Each simulated scenario incorporates a designated task route (as depicted in Fig. \ref{fig:simu_example_demos}) to evaluate the planning performance. The proposed method and baselines are executed to guide the AV along the provided task route. Specifically, the path generation method described in Sect. \ref{subsect:frenet representaion} is employed to facilitate the AV's merging onto the provided route. 
To ensure a valid and fair comparison of the final planning performance, lane change maneuvers are not permitted during the experiments. 
In cases where the implemented algorithm fails to find a valid solution, the AV will perform a stop behavior along the generated path, decelerating at a rate of $-4.0,m/s^2$. Regarding the planned trajectory, the tracking module utilizes the next time step's state in the planned trajectory as the AV's subsequent state, assuming perfect tracking by the AV. This reduces the influence of the redundant tracking module during the evaluation of planning performance.

\subsubsection{Prediction}

The prediction network \cite{deo2022-pgp-multimodal} is trained using feature data extracted from the closed-loop simulation itself, excluding the ego vehicle's involvement. The validation results are outlined in Table \ref{table:pred_results}, where values within the "Pct." column represent the percentage of training data employed during the training phase. This allows us to evaluate the influence of prediction performance on the subsequent planning task, a subject explored in the upcoming sections. 

Given that PGP \cite{deo2022-pgp-multimodal} is a multi-modal prediction method intended for a single agent, prediction calculations are carried out separately for each traffic agent at every simulation step. The K most probable prediction outcomes for each agent are directly incorporated into the planning process for collision avoidance or other pertinent calculations, without utilizing the likelihood information. Furthermore, the prediction horizon in the experiments is set to 6.0 seconds, with a 0.5-second interval. As the default option, we utilize the best prediction module, which is trained using all available training data.

\subsubsection{Auxiliary Functions}
To demonstrate the effectiveness of the proposed methodology, two auxiliary functions are implemented within the planning framework. The first function, named RP (Rear Predictions), preserves the prediction results from rear agents of the AV if enabled, and ignores rear predictions if disabled. Another function, IRD (Initial Relation Determination), corresponds to the content discussed in Section \ref{subsect:relation_deter}. When IRD is disabled, the interaction relation will be updated solely during the planning process. 

\vspace{-0.5em}
\subsection{Qualitative Results \label{subsect:experi_qualti_result}}

\begin{figure*}[tb!]
    \vspace{8pt} 
    \centering
    \subfloat[]{
        \includegraphics[width=1.1in]{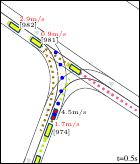}
        \includegraphics[width=1.1in]{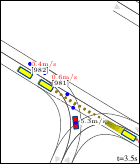}
        \includegraphics[width=1.1in]{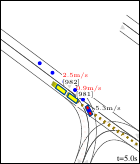}
        \includegraphics[width=1.1in]{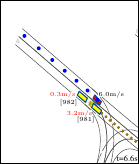}
        \includegraphics[width=1.1in]{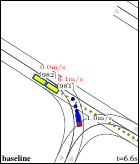}
        \label{subfig:quali_demo1}
        }\vspace{-0.5em}

    \subfloat[]{
        \includegraphics[width=1.1in]{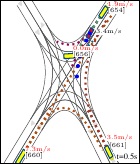}
        \includegraphics[width=1.1in]{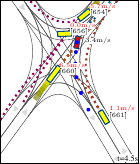}
        \includegraphics[width=1.1in]{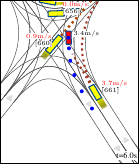}
        \includegraphics[width=1.1in]{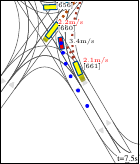}
        \includegraphics[width=1.1in]{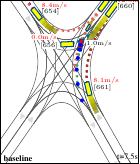}

        \label{subfig:quali_demo2}}

    \caption{Illustration of interaction results, where the AV is the red block. Other agents are depicted in yellow with their indexes in square brackets and their speeds shown in red (m/s). Additionally, the AV's speed is depicted in black, and its trajectory is represented by blue points with 1-second intervals.}
    \label{fig:simu_demos}
\vspace{-1.5em}
\end{figure*}

Fig. \ref{fig:simu_demos} presents two simulation outcomes that showcase the efficacy of the proposed algorithm. The final sub-figure labeled as ``baseline'' in each row displays the simulation result of the baseline, referred to as the Contingency planner \cite{cui2021lookout}, which will be introduced in Sect. \ref{subsect:experi_quanti_result}. The experiments are conducted with a prediction mode of $\text{K} = 3$.
\textbf{Fig. \ref{subfig:quali_demo1}}:
The AV initiates a left-turn with a speed of $4.5 \text{m/s}$ at 0.5 seconds, while potentially encountering interactions with agents 981 and 982. The simulation results demonstrate that the proposed method efficiently guides the AV through the conflict zone at 6.6 seconds. In contrast, the baseline makes limited progress at that time, with agent 981 still awaiting its way before the junction. This is due to the complexity of overtaking while maintaining a substantial safety time gap with the incoming prediction results from agents 981 and 982. Such a task is not easy and can result in a relatively conservative solution for the baseline method.
\textbf{Fig. \ref{subfig:quali_demo2}}:
In this simulation, unlike the baseline, which prompts the AV to slow down and yield to agent 660, the proposed method exhibits an interactive navigation capability that keeps its speed to traverse the conflict area. Simultaneously, agent 660 responds by decelerating to 0.9 m/s at 6.0 seconds and subsequently accelerating to 2.2 m/s by 7.5 seconds. This outcome can be attributed to the utilization of both (\ref{equation:relation_init_deter}) and (\ref{equation:influ_constraint_v2}) in our method, which together provide a broader solution space for the AV to overtake agent 660.

\vspace{-0.5em}
\subsection{Quantitative Results \label{subsect:experi_quanti_result}}

\renewcommand{\arraystretch}{1.0} 
\begin{table*}[t!]
\caption{Algorithm Comparisons (232 Scenarios)}
\centering
\vspace{-0.75em}
\begin{threeparttable} 
\begin{tabular}{p{2.5mm}|l*{3}c|*{11}c}
\Xhline{2\arrayrulewidth}
& & \multicolumn{3}{c}{Functions} & \multicolumn{3}{c}{Relation judgment\tnote{a}} & & \multicolumn{6}{c}{Driving performances}  \\ \cline{3-5} \cline{6-8} \cline{10-15}
ID\tnote{b} & Method & Pred-K & RP & IRD & \centering{$\underline{a}_i$} & \centering{$c_{f1}$} & \centering{$c_{f2}$} & \centering{$\underline{a}_i$} & DIST$\uparrow$ & FR$\downarrow$ & JERK$\downarrow$ & RC$\downarrow$ & CT$\downarrow$ & RCT$\downarrow$ \\ \hline 
A0 & CA & 1 & $\times$ & $\times$ & N/A & N/A & N/A & N/A & 47.92 & 6.95\%  & 4.22 & \textbf{0.628} & \textbf{15} & 8 \\ 
A1 & CA & 1 & $\checkmark$ & $\times$ & N/A & N/A & N/A & N/A & 45.68 & 10.16\%  & 5.82 & 0.630 & \textbf{15} & 11 \\
A2 & IR-Pred & 1 & $\times$ & $\times$ & N/A & N/A & N/A & N/A & 50.22 & 5.21\%  & 3.68 & 0.662 & 16 & \textbf{7} \\ 
A3 & IR-Pred & 1 & $\checkmark$ & $\times$ & N/A & N/A & N/A & -15.0 & 49.10 & 7.86\%  & 5.20 & 0.654 & \textbf{15} & 12 \\ 
A4 & IR-Pred & 1 & $\checkmark$ & $\checkmark$ & N/A & N/A & N/A & -15.0 & 49.98 & 5.54\% & 3.83 & 0.667 & 16 & 9 \\ 
\rowcolor[HTML]{E7E6E6}
A5 & IR-Influ & 1 & $\checkmark$ & $\checkmark$ & -0.01 & 1.0 & 3.0 & -15.0 & \textbf{51.55} & \textbf{4.75\%}  & \textbf{3.12} & 0.679 & \textbf{15} & 9 \\
\hline
A6 & IR-Pred & 3 & $\checkmark$ & $\checkmark$ & N/A & N/A & N/A & -15.0 & 43.59 & 8.52\%  & 6.23 & \textbf{0.600} & 15 & 8 \\ 
A7 & LS\cite{pan2020safe} & 3 & $\checkmark$ & $\checkmark$ & N/A & N/A & N/A & -15.0 & 47.07 & 7.29\%  & 4.64 & 0.652 & 16 & 7 \\
A8 & Conti\cite{cui2021lookout} & 3 & $\times$ & $\checkmark$ & N/A & N/A & N/A & -15.0 & 47.63 & \textbf{5.68\%} & 3.89 & 0.682 & 17 & 7 \\
\rowcolor[HTML]{E7E6E6}
A9 & IR-Influ & 3 & $\checkmark$ & $\checkmark$ & -0.01 & 1.0 & 3.0 & -15.0 & \textbf{48.10} & 5.83\%  & \textbf{3.88} & 0.656 & \textbf{14} & \textbf{6} \\
\Xhline{2\arrayrulewidth}
\end{tabular}

\smallskip
\scriptsize
\begin{tablenotes}
\RaggedRight
\item[a] parameters used in the (\ref{equation:irule_judge_relation}) during planning (Algo. \ref{alg:interaction_edge_checking}, Line 4), while another $\underline{a}_i$ is used in constraint checking (Algo. \ref{alg:interaction_edge_checking}, Line 7). \\
\item[b] indexes divide compared methods into several blocks for comparison, where \textbf{bold} values are the best results at each block.
\end{tablenotes}
\end{threeparttable}
\label{table:main_experiments}
\vspace{-1.5em}
\end{table*}
\renewcommand{\arraystretch}{1.0}

\subsubsection{Metrics}
To verify that the effectiveness of our proposed planning framework and modules, we conduct validation experiments with comprehensive settings. 
We compare algorithms using the following metrics: \textbf{DIST}: The average completion distance (in meters) of the AV along the given route in each scenario. \textbf{FR}: The fail rate of the algorithm used, reflects the algorithm's robustness in navigation performance. \textbf{JERK}: The average jerk cost defined by ($j^2 \cdot dt$), which reflects the planned trajectory's smoothness. In the equation, $dt$ is the interval time, which is set to 0.1 seconds. \textbf{RC}: Reaction cost of other traffic agents, which is defined as the average deceleration efforts ($a^2 \cdot dt$) of nearby agents within a 40-meter range. We define the reaction cost to quantify the planning performance of the proposed algorithm because, for an interactive planning method, the reactions of other traffic agents with the AV are taken into account during planning.
\textbf{CT}: The total number of valid collision times \cite{caesar2021nuplan} experienced by the AV, excluding collisions at the rear and collisions with agents when the AV is stationary. \textbf{RCT}: The collision times at the rear of the AV, a metric often overlooked in related studies, are introduced in our experiments to provide a more comprehensive assessment of experimental outcomes.

\subsubsection{Benchmarks}
Several baselines are utilized to validate the effectiveness of the proposed method. All of these methods are implemented using the same search framework discussed in Sect. \ref{subsect:s-t planning}, and they share a common set of parameters as specified in Table \ref{table:common_parameter_values}. \textbf{CA}: A traditional collision avoidance search method in which the spawned child does not consider its relations with other agents' prediction results but only avoids spatio-temporal intersections. We introduce this benchmark to show the effectiveness of interaction relation modeling in planning. \textbf{IR}-F: The proposed framework with the "F" formulation for determining interaction relations during the planning stage (Algo. \ref{alg:interaction_edge_checking}, Line 4). In the experiments, two relation determination functions, \textbf{IR-Pred} (corresponding to (\ref{equation:pred_judge_relation})) and \textbf{IR-Influ} (corresponding to (\ref{equation:irule_judge_relation})), are compared and discussed. \textbf{LS}: An optimization-based method relying on long-short-term distinguished prediction results. In this baseline, the most probable prediction result is preserved for the full horizon (long-term), while others are used with a shorter horizon (2.0 seconds) during planning. We implement this method in a search-based framework to demonstrate that interaction modeling benefits planning performance under multi-modal prediction results. \textbf{Conti}: Another method, referred to as the contingency planner \cite{cui2021lookout}, is introduced for comparison with our approach. This method devises a short-term trajectory (within 3.0 seconds) that ensures safety across all potential future scenarios and enables the creation of appropriate contingency plans for each potential realization (beyond 3.0 seconds).

\subsubsection{Validations}

Experimental results are shown in Table \ref{table:main_experiments}. 
Regarding the influence relation, as mentioned in Sect. \ref{subsubsect:rela_judge_dur_plan}, the agent's minimum response acceleration $\underline{a}_i$ has different values during the relation determination (\ref{equation:irule_judge_relation}) and constraint checking (\ref{equation:influ_constraint_v2}).
From A0-A2-A5 and the comparisons spanning from A6 to A9, it is evident that the proposed method excels over other benchmarks in terms of DIST, JERK, and CT metrics. In the FR metric, our method achieves the best performance among single-modal prediction outcomes, while securing the second-best yet comparable performance in the context of 3-modal prediction results. The observed effectiveness can be attributed to our incorporation of interaction modeling into the planning process. As anticipated, our method performs less favorably in the RC metric, as it is inherently designed to influence the motion of other agents, leading to more frequent instances of deceleration response. 

Functionalities of modules in our approach can be reflected in several aspects.
\textbf{Framework validation}: A0-A2 and A1-A3 pairs of comparisons prove that the proposed framework's incorporation of the interaction relation in planning enables a more robust interaction performance (lower FR and JERK metrics) when compared to traditional methods without interaction relation modeling. 
\textbf{IRD validation}: The A3-A4 comparison validates the effectiveness of the IRD module, which leads to improvements in nearly all planning metrics. 
\textbf{Validation of relation judgement formulations}: The comparison between A4 and A5 demonstrates the effectiveness of the relation determination function (\ref{equation:irule_judge_relation}) in contrast to (\ref{equation:pred_judge_relation}), resulting in varying degrees of improvement for the DIST, FR, and RC metrics.
\textbf{Processing rear predictions}: As suggested by the A2-A3-A4 comparisons, omitting the prediction results from the rear appears to contribute the most significant enhancement in all performance metrics, including the RCT. However, this observation might be owing to the fact that most of our experiments are conducted without a lane-change scenario. Instead of excluding them, the proposed IRD module transforms the rear prediction results into a constraint formulation controlled by $\underline{a}_i$, resulting in a similar driving performance (as observed in the A2-A4 comparison). This adaptability allows for configuration in various situations when applied, and more details will be discussed in Sect. \ref{subsubsect:exp_paramtest}.

\begin{figure}[!t]
    \centering
    \includegraphics[width=0.7\linewidth]{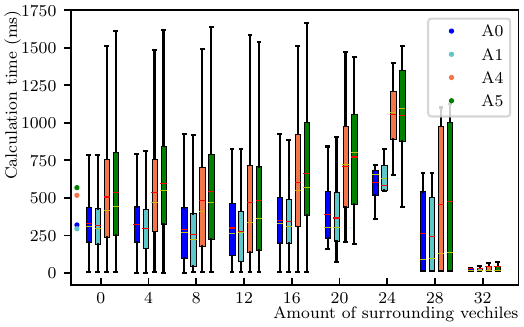} \vspace{-0.5em}
    \caption{This illustrates the calculation time during the planning phase of the simulation. The horizontal axis represents the number of vehicles within the AV's observation window. The colored dots indicate the average calculation time for each method. Within a box (Gaussian distribution), the red line represents the mean value, and the yellow line represents the median value. }
    \label{fig:caculation_times}
\vspace{-1.5em}
\end{figure}

\subsubsection{Time Complexity \label{subsubsect:exp_time_complexity}} Fig.\ref{fig:caculation_times} illustrates the calculation times of experiments in Table \ref{table:main_experiments}. It is evident that the calculation times of methods that consider interaction relations (A4, A5) are approximately two to three times longer than those of the naive collision checking methods (A0, A1) at all times, as expected. Moreover, the calculation times appear to be unaffected by the number of surrounding vehicles, as increasing the number narrows the solution space. While the implemented algorithms do not currently achieve real-time performance, this is acceptable in the simulation context. Implementing them in a C++ backend can further enhance their efficiency.

\subsubsection{Influence of Prediction Results \label{subsubsect:exp_multi_modal_test}}

\renewcommand{\arraystretch}{1.0} 
\begin{table}[t!]
\caption{Tests under Multi-modal Prediction Results (76 Scenarios)}
\centering
\vspace{-0.75em}
\begin{threeparttable} 
\begin{tabular}{p{3.0mm}|p{10.5mm}p{5.0mm}|p{6.0mm}p{6.0mm}p{6.0mm}p{4.0mm}p{3.0mm}p{3.0mm}}
\Xhline{2\arrayrulewidth}
ID & Method & P-K & DIST$\uparrow$ & FR$\downarrow$ & JERK$\downarrow$ & RC$\downarrow$ & CT$\downarrow$ & RCT$\downarrow$ \\ \hline
B0 & IR-Pred & 1 & 41.81 & 4.97\%  & 4.14 & 0.600 & 3 & \textbf{3} \\ 
B1 & IR-Pred & 2 & 35.86 & 7.62\%  & 6.54 & 0.630 & 3 & 4 \\ 
B2 & IR-Pred & 3 & 31.48 & 9.54\%  & 8.42 & \textbf{0.591} & \textbf{2} & 4 \\ 
B3 & LS\cite{pan2020safe} & 1 & 41.81 & 4.97\%  & 4.14 & 0.600 & 3 & \textbf{3} \\
B4 & LS\cite{pan2020safe} & 2 & 38.35 & 6.91\%  & 5.92 & 0.594 & 3 & \textbf{3} \\
B5 & LS\cite{pan2020safe} & 3 & 36.25 & 7.99\%  & 5.87 & 0.599 & \textbf{2} & \textbf{3} \\
B6 & Conti\cite{cui2021lookout} & 1 & 43.11 & 3.75\%  & 3.18 & 0.650 & \textbf{2} & 4 \\
B7 & Conti\cite{cui2021lookout} & 2 & 39.56 & 5.21\%  & 4.47 & 0.633 & \textbf{2} & 4 \\
B8 & Conti\cite{cui2021lookout} & 3 & 36.77 & 5.92\%  & 4.92 & 0.644 & \textbf{2} & 4 \\
B9 & IR-Influ & 1 & \textbf{44.96} & \textbf{3.73\%}  & \textbf{2.89} & 0.655 & \textbf{2} & \textbf{3} \\ 
B10 & IR-Influ & 2 & 42.18 & 5.10\%  & 3.94 & 0.606 & \textbf{2} & \textbf{3} \\
B11 & IR-Influ & 3 & 40.28 & 5.69\%  & 4.61 & 0.634 & \textbf{2} & \textbf{3} \\
\Xhline{2\arrayrulewidth}
\end{tabular}

\smallskip
\scriptsize
\begin{tablenotes}
\RaggedRight
\item[] Other function and parameter settings are in line with A4 (IR-Pred), A5 (IR-Influ), A7 (LS), and A8 (Conti) in Table \ref{table:main_experiments}. \\
\end{tablenotes}
\end{threeparttable}
\label{table:multi_modal_experiments}
\vspace{-1.5em}
\end{table}
\renewcommand{\arraystretch}{1.0}

In complex urban environments, achieving accurate predictions of traffic agents' movements is challenging due to the inherent uncertainty of traffic behavior. Consequently, it is essential to investigate the impact of prediction modules on downstream planning performance. This paper includes two aspects of the study: one focuses on examining the effects of multi-modal prediction results, while the other investigates prediction accuracy.

The comparison results with different prediction modes (Pred-K, P-K in short) are displayed in Table \ref{table:multi_modal_experiments}. Upon analyzing these experiments, along with the A4-A6 comparisons in Table \ref{table:main_experiments}, it appears that an increased utilization of prediction results (P-K) does not significantly enhance planning safety (as indicated by the CT metric). Our speculation is that a single-modal prediction module might approximate multi-modal prediction results across consecutive frames. Although the elevation of P-K leads to a decrease in DIST, FR, and JERK, the proposed method consistently showcases the best performance across all P-K conditions. In addition, our method exhibits the least performance degradation, with DIST reduced to 41.08, FR to 5.77\%, JERK to 4.81, and even RC to 0.59.
outperforming all other benchmarks.

\begin{figure}[!t]
\centering

\subfloat[Pred-K=1 (A4, A5 settings)]{
    \includegraphics[height=1.5in]{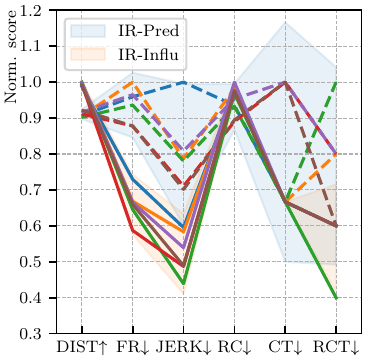}
    \label{subfig:exp_predtest_p1}}
\subfloat[Pred-K=3 (A8, A9 settings)]{
    \includegraphics[height=1.5in]{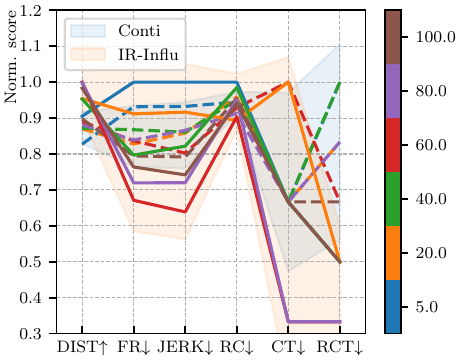}
    \label{subfig:exp_predtest_p3}}

\caption{Experiments (76 scenarios) under different prediction networks, where solid lines represent the performance of the IR-Influ, and dashed lines are the baselines. The color and its value on the bar indicate the corresponding Pct. value of the prediction module (Table \ref{table:pred_results}) utilized in the experiment, and the metrics are normalized by (a): [45.35, 5.66\%, 5.90, 0.671, 3, 5], and (b): [40.98, 7.45\%, 6.21, 0.674, 3, 6], respectively.}
\label{fig:exp_prednoise_test}
\vspace{-1.5em}
\end{figure}

Considering the diverse prediction modules outlined in Table \ref{table:pred_results}, we now move on to comparing the performances of planning methods under different prediction accuracies. By planning with prediction modules trained on varying amounts of training data, we can simulate the long-tail training effects on the prediction module and the downstream planning process. This is because, in practice, not all scenarios can be fully trained.
The experimental results are depicted in Fig. \ref{fig:exp_prednoise_test}, where subfigures (a) and (b) indicate that higher prediction accuracy can enhance planning performance, but exceeding a certain level of accuracy does not necessarily yield further improvement. In (a), our method demonstrates significantly more robust and superior performance in all metrics except for RC, compared to the benchmark. In (b), our method exhibits better performance when the Pct. value is greater than 50.0\%. This suggests that our method's robustness is somewhat less than that of Conti \cite{cui2021lookout} under multi-modal prediction results.

\subsubsection{Influences of Algorithm Coefficients \label{subsubsect:exp_paramtest}}

In Table \ref{table:param_experiments1} and \ref{table:param_experiments2}, we performed several experiments to evaluate how coefficients from (\ref{equation:influ_constraint_v2}) and (\ref{equation:influ_constraint2}) impact the navigation performance.

We first validate the effectiveness of the $\underline{a}_i$ value used in the processes of relation determination and constraint checking (\ref{equation:influ_constraint_v2}) respectively. In Table \ref{table:param_experiments1}, we set up group C experiments under $[c_{f1}, c_{f2}] = [-0.5, 0.0]$ conditions because they share similar functions that define the safety boundary in establishing an influence relation. The experiments demonstrate that as the value of $\underline{a}_i$ increases, the DIST, FR, and JERK metrics all decrease, while the RC and CT metrics show improvement. In our study, we distinguish values of $\underline{a}_i$ as $[-0.01, -15.0]$ to ensure that the AV initially behaves as a reactor and then exhibits influencer characteristics.
Next, we discuss the experiments with different gap time coefficient $c_{f1}$ in the E group experiments of Table \ref{table:param_experiments2}. When $c_{f1}$ increases, fewer interactions are distinguished as influence relations during planning, resulting in a monotonic decrease in DIST and RC metrics, and a monotonous increase in JERK metric. In comparison, the FR and CT achieve their relatively low values at 1.0. This indicates that an appropriate $c_{f1}$ value can achieve a balance between smoothness (FR and JERK) and safety (CT), while a too large value leads to a purely reactive planner (E6's performance equals B0's), and a too small value (e.g., -0.5) results in an aggressive (high DIST and low JERK) and unsafe performance (CT equals 7). Lastly, performances under different $c_{f2}$ coefficients are compared in group F, as shown in Table \ref{table:param_experiments2}. The $c_{f2}$ is designed to encourage establishing influence relations at relatively higher speed conditions, making the influence relations more stable. Evidence is presented in the FR metric, which achieves its lowest value of 3.42\% at 1.0. However, as $c_{f2}$ increases from 1.0 to 5.0, it gradually discourages low-speed influence relations, resulting in slight increases in JERK and FR metrics, and steady decreases in the RC metric. In contrast, DIST is only slightly changed, implying that the $c_{f2}$ coefficient has little impact on the planning performance's level of aggressiveness.

\renewcommand{\arraystretch}{1.0} 
\begin{table}[t!]
\caption{Coefficient Test Results 1 (76 Scenarios)}
\centering
\vspace{-0.75em}
\begin{threeparttable} 
\begin{tabular}{p{2.5mm}|p{6.0mm}p{6.0mm}|p{6.0mm}p{6.0mm}p{6.0mm}p{6.0mm}p{4.0mm}p{4.0mm}}
\Xhline{2\arrayrulewidth}
ID & \centering{$\underline{a}_i$}\tnote{1} & \centering{$\underline{a}_{i}$} & DIST$\uparrow$ & FR$\downarrow$ & JERK$\downarrow$ & RC$\downarrow$ & CT$\downarrow$ & RCT$\downarrow$ \\ \hline
C0 & -15.0 & -15.0 & \textbf{46.71} & \textbf{3.52\%}  & \textbf{2.13} & 0.690 & 4 & \textbf{3} \\
C1 & -6.0 & -15.0 & 46.53 & 3.63\%  & 2.16 & 0.682 & 4 & \textbf{3} \\
C2 & -1.5 & -15.0 & 46.32 & 3.85\%  & 2.27 & \textbf{0.670} & 4 & \textbf{3} \\
C3 & -0.01 & -15.0 & 45.94 & 3.93\%  & 2.40 & 0.672 & \textbf{3} & \textbf{3} \\
\hline
D0 & -0.01 & -15.0 & 45.46 & \textbf{3.44\%}  & 2.56 & 0.640 & 3 & \textbf{2} \\
D1 & -0.01 & -6.0 & \textbf{45.47} & 3.47\%  & \textbf{2.55} & 0.643 & 3 & \textbf{2} \\
D2 & -0.01 & -1.5 & 44.86 & 3.98\%  & 2.85 & 0.642 & 3 & 3 \\
D3 & -0.01 & -0.01 & 43.26 & 5.16\%  & 3.77 & \textbf{0.620} & \textbf{2} & 3 \\

\Xhline{2\arrayrulewidth}
\end{tabular}

\smallskip
\scriptsize
\begin{tablenotes}
\RaggedRight
\item[1] parameter used in the relation determination (\ref{equation:irule_judge_relation}) during planning, while another is used in constraint checking. Other function and parameter settings are consistent with A5 in Table \ref{table:main_experiments}, except for group C ($[c_{f1},c_{f2}]=[-0.5,0.0]$) and group D ($[c_{f1},c_{f2}]=[0.5,1.0]$). \\
\end{tablenotes}
\end{threeparttable}
\label{table:param_experiments1}
\vspace{-1.0em}
\end{table}
\renewcommand{\arraystretch}{1.0}

\renewcommand{\arraystretch}{1.0} 
\begin{table}[t!]
\caption{Coefficient Test Results 2 (76 Scenarios)}
\centering
\vspace{-0.75em}
\begin{threeparttable} 
\begin{tabular}{p{2.5mm}|p{6.0mm}p{6.0mm}|p{6.0mm}p{6.0mm}p{6.0mm}p{6.0mm}p{4.0mm}p{4.0mm}}
\Xhline{2\arrayrulewidth}
ID & \centering{$c_{f1}$} & \centering{$c_{f2}$} & DIST$\uparrow$ & FR$\downarrow$ & JERK$\downarrow$ & RC$\downarrow$ & CT$\downarrow$ & RCT$\downarrow$ \\ \hline
E0 & -0.5 & 0.0 & \textbf{46.99} & \textbf{3.54\%}  & \textbf{2.18} & 0.730 & 6 & 
\textbf{3} \\ 
E1 & 0.01 & 0.0 & 45.00 & 3.77\%  & 2.63 & 0.667 & 3 & \textbf{3} \\ 
E2 & 0.5 & 0.0 & 45.53 & 3.72\%  & 2.71 & 0.659 & 3 & 3 \\
E3 & 1.0 & 0.0 & 44.91 & 3.72\%  & 2.87 & 0.646 & \textbf{2} & \textbf{3} \\ 
E4 & 2.0 & 0.0 & 43.67 & 3.83\%  & 3.17 & 0.646 & 4 & \textbf{3} \\ 
E5 & 4.0 & 0.0 & 42.00 & 4.74\%  & 4.04 & 0.609 & 3 & 4 \\ 
E6 & 6.0 & 0.0 & 41.81 & 4.97\%  & 4.14 & \textbf{0.600} & 3 & \textbf{3} \\ 
\hline
F0 & 0.5 & 0.01 & 45.32 & 3.55\%  & 2.62 & 0.646 & \textbf{3} & 3 \\ 
F1 & 0.5 & 0.5 & 45.19 & 3.57\%  & \textbf{2.59} & \textbf{0.641} & \textbf{3} & 3 \\ 
F2 & 0.5 & 1.0 & 45.27 & \textbf{3.47\%}  & 2.62 & \textbf{0.641} & \textbf{3} & \textbf{2} \\ 
F3 & 0.5 & 2.0 & 45.39 & 3.55\%  & 2.70 & 0.650 & \textbf{3} & 3 \\ 
F4 & 0.5 & 3.0 & \textbf{45.53} & 3.72\%  & 2.71 & 0.659 & \textbf{3} & 3 \\ 
F5 & 0.5 & 5.0 & 45.34 & 3.87\%  & 3.08 & 0.645 & \textbf{3} & 3 \\ 
\Xhline{2\arrayrulewidth}
\end{tabular}

\smallskip
\scriptsize
\begin{tablenotes}
\RaggedRight
\item[] For better comparisons, all function and parameter settings are kept the same as those in A5 of Table \ref{table:main_experiments} with the exception of $\underline{a}_i(-15.0)$.
\end{tablenotes}
\end{threeparttable}
\label{table:param_experiments2}
\vspace{-1.5em}
\end{table}
\renewcommand{\arraystretch}{1.0}

\subsubsection{Experiments under Different Aggressiveness Levels}

\begin{figure}[!t]
    \centering
    \subfloat[Pred-K=1 (A4, A5 settings)]{
        \includegraphics[height=1.5in]{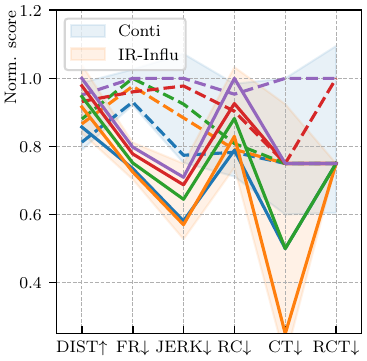}
        \label{subfig:exp_coeftest_p1}}
    \subfloat[Pred-K=3 (A8, A9 settings)]{
        \includegraphics[height=1.5in]{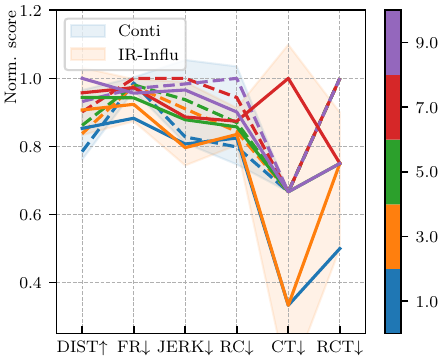}
        \label{subfig:exp_coeftest_p3}}

    \caption{Experiments (76 scenarios) under different aggressiveness levels, where solid lines represent the performance of the IR-Influ, and dashed lines are the baselines. The color and its value on the bar represent the value of $w_v$ utilized in the algorithm, and the metrics are normalized by (a): [47.48, 4.97\%, 4.48, 0.742, 4, 4], and (b): [42.67, 6.04\%, 5.25, 0.740, 3, 4], respectively.
    }
    \label{fig:exp_aggressiveness_test}
\vspace{-1.5em}
\end{figure}

To further validate the effectiveness of our method, we compare the performances of the algorithms at different levels of aggressiveness, as indicated by the weight parameter $w_v$ in Table \ref{table:common_parameter_values}. 
Fig. \ref{fig:exp_aggressiveness_test} illustrates the experimental results, showing that as the $w_v$ increases from 1.0 to 9.0, the DIST and JERK metrics exhibit a monotonic increase. In (a), our method consistently outperforms across all metrics except for RC, showcasing consistent changes in DIST, FR, JERK, and RC metrics as $w_v$ increases. In (b), our method maintains its superiority over the baseline in DIST, FR, and JERK metrics, but exhibits a decline in CT performance when $w_v$ equals 7.0.

\section{Conclusion \label{sect:discussion_conclusion}}

In this study, we present an interactive planning framework for autonomous driving. Our approach introduces the concept of an ``interaction zone'' to model traffic interactions. These interactions are mathematically formulated as influencer and reactor relations, with the reactor relation including overtaking and yielding interactions. 
The integration of these elements into a spatio-temporal search framework, with continuous maintenance and updating of interaction relations, imposes constraints during forward expansion, enabling the derivation of interaction-aware trajectories. 

Simulation results comparing our approach to benchmarks demonstrate several advantages:
i) The proposed method demonstrates enhanced effectiveness and robustness in planning under single-modal prediction, resulting in substantial improvements of 7.6\% in distance completeness and 31.7\% in the fail rate metrics. Furthermore, in three-modal scenarios, it exhibits a comparatively higher level of security with a reduction of approximately 17.6\% in collision times. ii) In our method, the interaction performance can be fine-tuned by explicitly defining coefficients, thereby enhancing the algorithm's scalability for application in novel scenarios. iii) Our method demonstrates a significant level of adaptability to diverse prediction accuracies and levels of aggressiveness. This is achieved through the incorporation of interaction modeling, which goes beyond solely relying on prediction outcomes and instead considers the potential reactions of other agents. 

Our future endeavors aim to explore and incorporate a wider range of interaction relations in the context of autonomous driving, extending beyond the conventional roles of ``influencers'' and ``reactors''. Moreover, we recognize that these relationships cannot solely be modeled based on the overlapping trajectories of vehicles. For instance, a vehicle's behavior may be influenced by the presence of pedestrian traveling nearby it. Additionally, an area for future study involves modeling the interaction relations between vehicles and pedestrians, as well as incorporating traffic rules related to traffic lights and crosswalks. These enhancements will significantly increase the adaptability of the planning method in various scenarios.

\section{Acknowlegement \label{sect:acknowlegement}}

The authors extend their sincere appreciation to Lotus Technology Ltd. for their financial support and valuable discussions with their staffs on topics related to this work.



\bibliographystyle{IEEEtran}
\bibliography{IEEEabrv,ref}



\vspace{-60pt}

\begin{IEEEbiography}[{\includegraphics[width=1.0in,height=1.2in,clip,keepaspectratio]{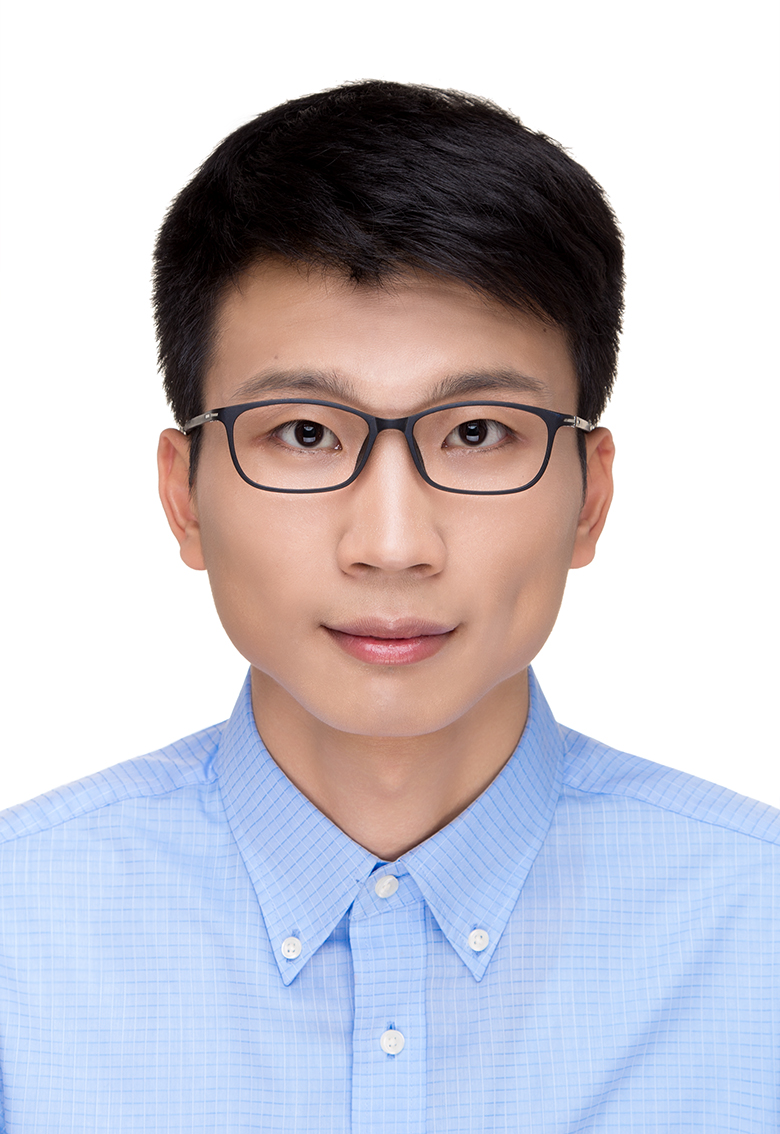}}]{Yingbing Chen}
(Student Member, IEEE) received the B.Sc. degree from Northwestern Polytechnical University, Xian, China, in 2015, and the the M.S. degree from Xiamen University, Xiamen, China, in 2018.  He is currently working toward the Ph.D. degree with the Division of Emerging Interdisciplinary Areas, The Hong Kong University of Science and Technology, HKSAR, China, supervised by Prof. M. Liu. His current research interests include machine learning, motion and behavioral planning for autonomous driving and robotics.
\end{IEEEbiography}
\vspace{-23pt}

\begin{IEEEbiography}[{\includegraphics[width=1.0in,height=1.2in,clip,keepaspectratio]{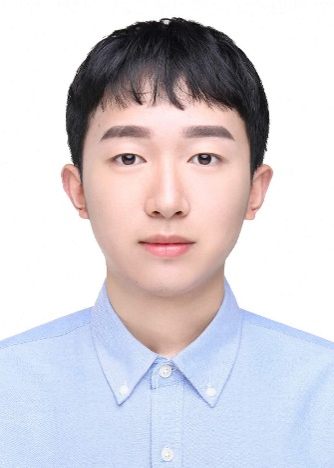}}]{Jie Cheng}
received the B.S. degree from Huazhong University of Science and Technology, Wuhan, China, in 2019. 
He is currently pursuing the Ph.D. degree in the Department of Electronic and Computer Engineering, the Hong Kong University of Science and Technology, HKSAR, China, supervised by Prof. Ming Liu.
His research mainly focuses on motion planning and motion forecasting for autonomous driving.
\end{IEEEbiography}
\vspace{-23pt}

\begin{IEEEbiography}[{\includegraphics[width=1.0in,height=1.2in,clip,keepaspectratio]{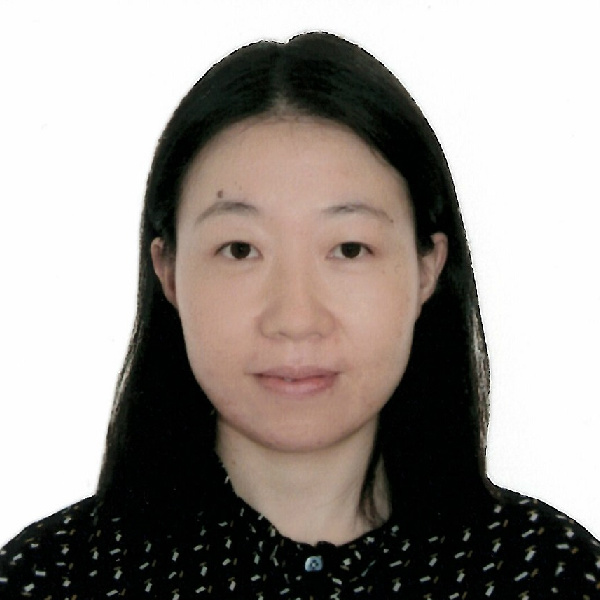}}]{Lu Gan}
    Lu, GAN received both her bachelor and master degreee from Nanjing University of Aeronautics and Astronautics. After that, she worked in Nanyang Technological University as a research associate. Now she is a PhD student in Hong Kong University od Science and Technology. Her research interests include autonomous driving, deep learning and uncertainty-aware motion prediction and planning. 
    \end{IEEEbiography}
    \vspace{-23pt}

\begin{IEEEbiography}[{\includegraphics[width=1.0in,height=1.2in,clip,keepaspectratio]{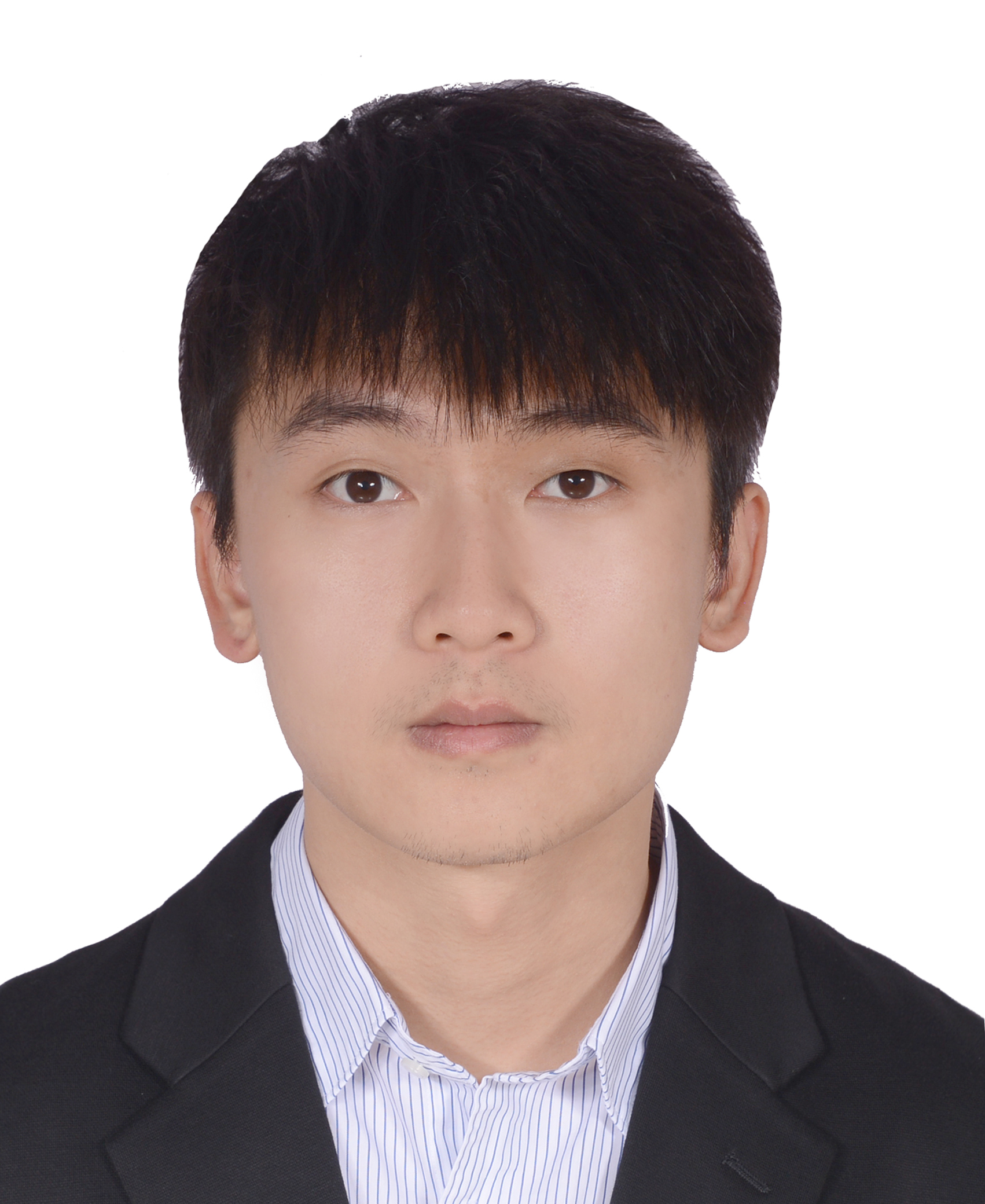}}]{Sheng Wang}
(Student Member, IEEE) received the B.S. degree from Harbin Institute of
Technology, Harbin, China, in 2018, and the the M.S. degree from École Centrale de Nantes, Nantes, France, in 2020. He is currently working toward the Ph.D. degree 
with the Division of Emerging Interdisciplinary Areas, The Hong Kong University of Science and Technology, supervised by Prof. Ming Liu. His research interests include imitation learning, reinforcement learning, trajectory prediction and planning.
\end{IEEEbiography}
\vspace{-23pt}

\begin{IEEEbiography}[{\includegraphics[width=1.0in,height=1.2in,clip,keepaspectratio]{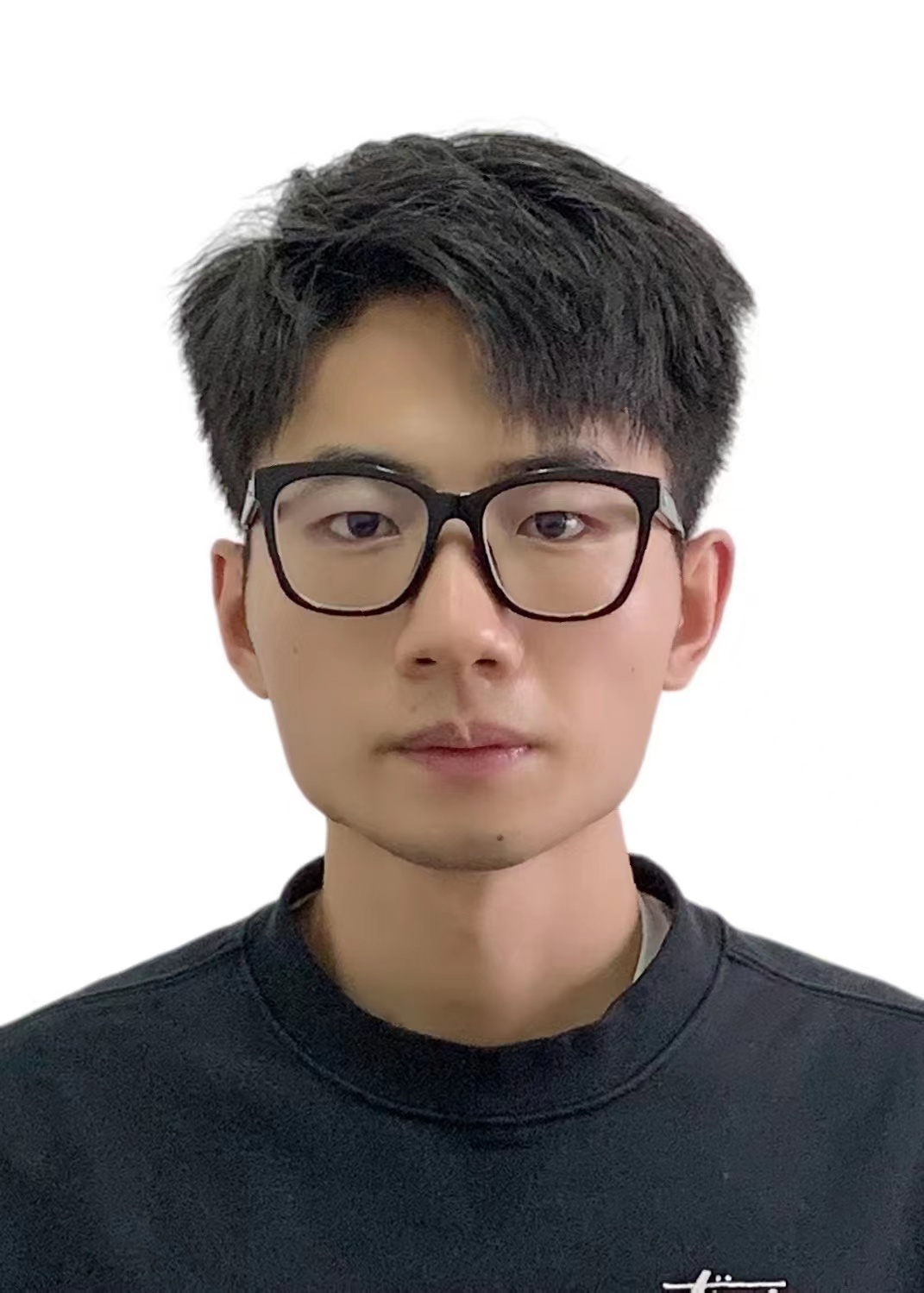}}]{Hongji Liu}
is a Ph.D. candidate in robotics and autonomous systems at the Intelligent Autonomous Driving Center (IADC), The Hong Kong University of Science and Technology, under the supervision of Prof. Ming Liu and Prof. Qifeng Chen. Before that, he received his bachelor's degree in Software Engineering at Sun Yat-sen University. His research interests mainly include high-definition map generation, topological navigation, and SLAM (Simultaneous Localization and Mapping).
\end{IEEEbiography}
\vspace{-23pt}

\begin{IEEEbiography}[{\includegraphics[width=1.0in,height=1.2in,clip,keepaspectratio]{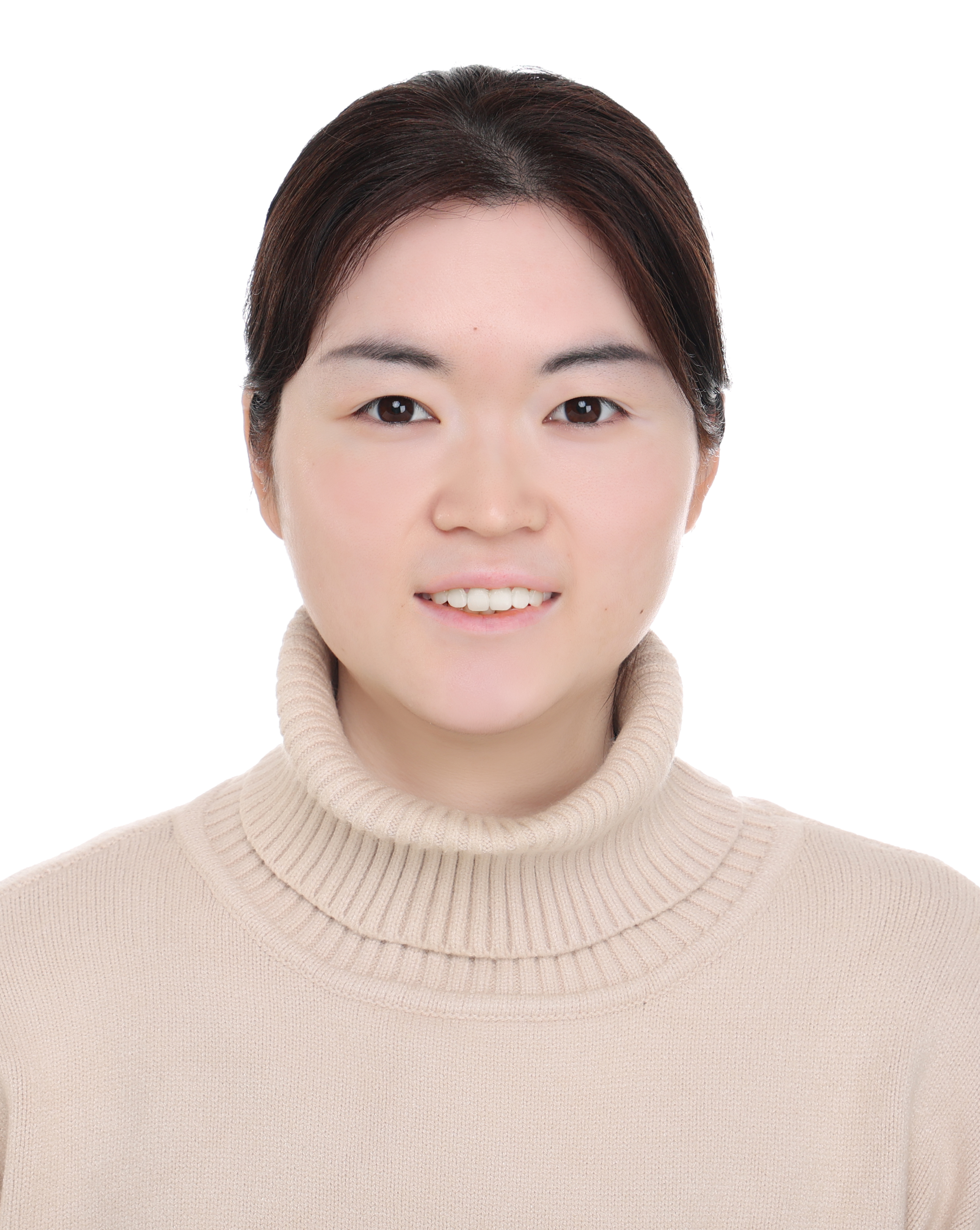}}]{Xiaodong Mei}
(Student Member, IEEE) received the B.S. degree from University of Chinese Academy of Sciences, Beijing, China, in 2018.  She is currently pursuing the Ph.D. degree in the Department of Computer Science and Engineering, The Hong Kong University of Science and Technology, HKSAR, China, supervised by
Prof. Ming Liu. Her current research interests include motion forecasting and planning for autonomous driving and robotics.
\end{IEEEbiography}
\vspace{-22pt}

\begin{IEEEbiography}[{\includegraphics[width=1.0in,height=1.2in,clip,keepaspectratio]{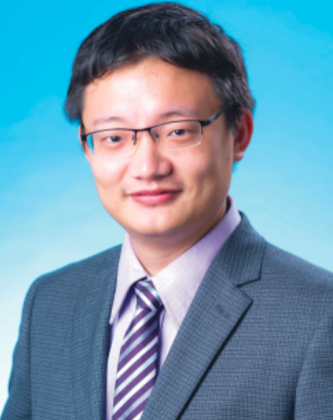}}]{Ming Liu}
(Senior Member, IEEE) received the B.A. degree in automation from Tongji University, Shanghai, China, in 2005, and the Ph.D. degree from the Department of Mechanical and Process Engineering, ETH Zurich, Zurich, Switzerland, in 2013, supervised by Prof. R. Siegwart. He is currently with the Department of Electronic and Computer Engineering and Computer Science and Engineering Department, Robotics Institute, The Hong Kong University of Science and Technology, Hong Kong, as an Associate Professor. He is currently also the Chairman with Shenzhen Unity Drive Inc., China. His research interests include dynamic environment modelling, deep-learning for robotics, 3-D mapping, machine learning, and visual control. He has authored and co-authored many popular papers in top robotics journals including IEEE TRANSACTIONS ON ROBOTICS, and IEEE ROBOTICS AND AUTOMATION MAGAZINE. 

\end{IEEEbiography}


\end{document}